\documentclass[lettersize,journal]{IEEEtran}

\usepackage{amsmath,amsfonts}
\usepackage{algorithmic}
\usepackage{array}
\usepackage{textcomp}
\usepackage{stfloats}
\usepackage{url}
\usepackage{verbatim}
\usepackage{color}
\usepackage{float}
\usepackage{bm}
\usepackage{amssymb}
\usepackage{threeparttable}
\usepackage{multirow}
\usepackage{eqnarray}
\usepackage{color}
\usepackage{tablefootnote}

\usepackage{amsmath} 
\usepackage{amssymb}  
\usepackage{color}
\usepackage[english]{babel}
\usepackage{amsthm}
\usepackage{breqn}
\usepackage[super]{nth}
\usepackage{url}
\usepackage{siunitx}
\usepackage{makecell}
\usepackage{microtype}
\usepackage{booktabs}
\usepackage{stfloats}
\usepackage{threeparttable} 
\usepackage{multirow}
\theoremstyle{definition}

\usepackage{verbatim}

\usepackage[utf8]{inputenc}
\usepackage{graphicx}
\usepackage{enumitem}
\usepackage[font=footnotesize]{caption}
\usepackage[font=footnotesize]{subcaption}
\usepackage{float}
\usepackage{bm}
\usepackage{amssymb}
\usepackage{threeparttable}
\usepackage{multirow}
\usepackage{eqnarray}
\usepackage{color}
\usepackage{tablefootnote}
\usepackage[table,xcdraw]{xcolor}
\usepackage[ruled,vlined]{algorithm2e}

\definecolor{blue}{rgb}{0, 0, 0}
\definecolor{linkblue}{rgb}{0, 0.19, 0.32}
\usepackage[colorlinks, linkcolor=black]{hyperref}
\hypersetup{
    colorlinks=true, 
    urlcolor=linkblue
    }

\newcommand{\blue}{\textcolor{blue}}
\def\BibTeX{{\rm B\kern-.05em{\sc i\kern-.025em b}\kern-.08em
    T\kern-.1667em\lower.7ex\hbox{E}\kern-.125emX}}
\usepackage{balance}

\makeatletter
\def\fnum@figure{Fig.~\thefigure}
\makeatother

\begin{document}
\title{Arm-Aware Guided Dexterous Grasp Generation with Arm-Agnostic Grasp Models}
\author{Yongyi Jia$^{\dagger}$, Yongpeng Jiang$^{\dagger}$, Kangchen Lv, Yi Ren, Mingrui Yu$^{*}$, and Xiang Li$^{*}$
\thanks{$^{\dagger}$Equal contribution. 
$^{*}$ Co-corresponding authors.
This work was supported in part by the Brain Science and Brain-like Intelligence Technology-National Science and Technology Major Project under Grant 2021ZD0201404, in part by the National Natural Science Foundation of China under Grant 62461160307, in part by the Fundamental and Interdisciplinary Disciplines Breakthrough Plan
of the Ministry of Education of China under Grant JYB2025XDXM208, and in part by the BNRist project under Grant BNR2024TD03003. 
}}

\maketitle

\begin{abstract}
Dexterous grasp generation that considers arm-related constraints is crucial in real-world scenarios involving arm-environment collision avoidance, workspace boundary grasps, and consecutive grasping. Existing hand-centric grasp models, which primarily focus on the floating hand’s pose, are insufficient for such cases. Conventional arm-aware methods either rely on rejection sampling to discard infeasible samples or require retraining on arm-specific data, leading to low sample efficiency under adverse conditions or limited generalization across different robots and environments. To overcome these limitations, this letter presents an arm-aware dexterous grasp generation framework that leverages pretrained arm-agnostic grasp models while integrating arm and environmental information only at inference time. Specifically, we formulate arm-aware constrained grasp generation as a joint optimization of hand pose and arm configuration, and derive closed-form gradients for arm-related constraints. Assuming the hand pose distribution is represented by a diffusion model, we prove that gradient-based optimization is equivalent to guided diffusion sampling, steering near-feasible samples toward the feasible region.
Through comprehensive evaluation involving 10k objects across 6 scenarios, we demonstrate that the proposed framework generates feasible grasps in highly constrained settings with significantly higher probability, highlighting its advantages in real-world applications.
Supplementary materials and appendix are available at \texttt{\url{https://arm-aware-dexgrasp.github.io/}}.
\end{abstract}


\section{Introduction}
\IEEEPARstart{D}{exterous} grasp generation, which generates grasp poses for dexterous hands based on object and environment information, provides a target grasp configuration
for grasp execution \cite{Chen2024BODexSA, wei2024learning}, and serves as a prerequisite for the subsequent robotic manipulation \cite{Jiang2024ContactImplicitMP}.
%

%
Existing grasp generation methods predominantly employ a \textit{hand-centric} scheme, 
which focuses primarily on learning the distribution of a free-floating hand’s grasp poses, overlooking the robotic arm and critical environmental context, such as obstacles \cite{wei2024learning, Zhang2024DexGraspNet2L, Zhong2025DexGraspAT}.
However, focusing solely on the hand is often insufficient in practice. For example, when grasping in a constrained space,
the arm must avoid collisions with surrounding objects to ensure safety. Moreover, grasping objects near the boundary of the arm's reachable space constrains the wrist orientation to ensure the existence of inverse kinematics (IK) solutions. Additionally, minimizing arm motion is important for execution efficiency, particularly in tasks involving consecutive grasping and confined pick-and-place. Therefore, it is essential to incorporate arm-related constraints into the grasp generation process.
\begin{figure}[!tb]
    \centering
    \includegraphics[width=1\linewidth]{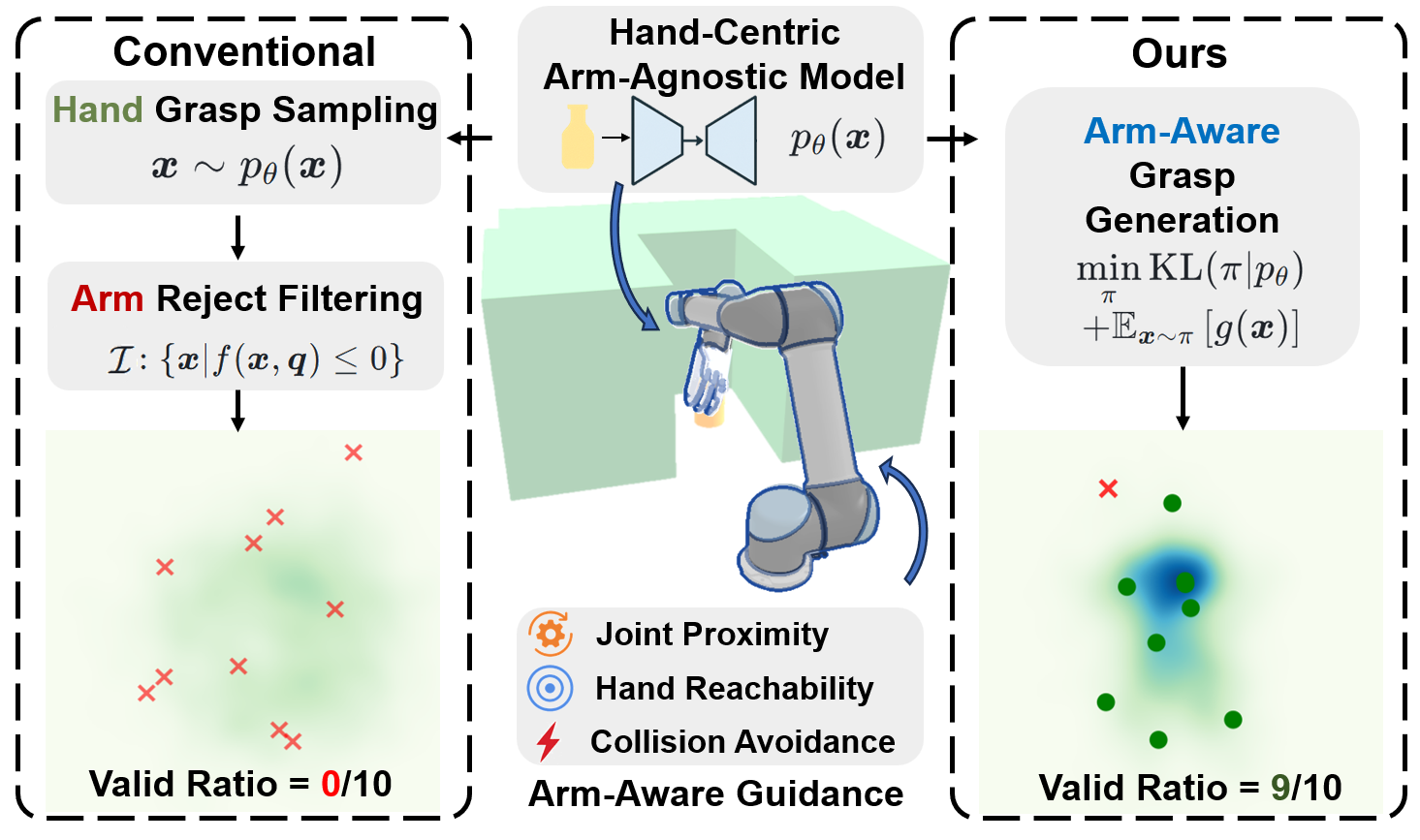}
    \vspace{-5pt}
    \caption{Motivation of arm-aware dexterous grasp generation. Real-world grasp execution requires considering arm-related constraints, yet conventional methods rely on rejection sampling to discard even near-feasible hand poses, leading to low efficiency. Our approach guides pretrained arm-agnostic models with these constraints at inference, greatly improving sampling efficiency.}
    \label{fig: head}
    \vspace{-15pt}
\end{figure}
%
%

%
To address this issue, a straightforward solution is \textit{rejection sampling}, where multiple candidate grasps are generated and those violating arm-related constraints are filtered out before execution.
Although this strategy is widely used in practice, it significantly reduces sampling efficiency—especially in highly constrained environments where the feasible region occupies a small volume relative to the full solution space, as many near-feasible samples are discarded without correction.
An alternative is to directly train a model on synthetic data that involves both the hand and arm, which already satisfies the constraints \cite{Chen2024BODexSA}.
However, this approach suffers from limited generalizability, as the grasp model tends to overfit to a specific arm and environment setup. Adapting to a different robotic arm or environment necessitates additional data synthesis and retraining the model, which is time-consuming.
%

%
%

%
In contrast, this letter proposes an \textit{arm-aware} dexterous grasp generation method (shown in Fig. \ref{fig: head}) by reformulating the problem as a joint optimization of grasp pose and arm configuration, utilizing only pre-trained arm-agnostic (i.e., hand-centric) grasp generation models.
Our approach enhances sampling efficiency by transforming near-feasible samples into feasible ones through inference-phase guidance with arm-related constraints.
Additionally, we express all arm-related constraints using closed-form equations and gradients, enabling our approach to be applicable to any robotic arms and environments, based on pre-modeled analytical arm kinematics and environment SDFs, without the need for costly data synthesis and re-training.
Assuming the arm-agnostic grasp distribution is expressed as a diffusion model, we demonstrate that solving the proposed optimization problem using a primal–dual (PD) method corresponds to guided grasp-pose denoising within the diffusion framework.
While incorporating guidance in the diffusion framework is common, designing guidance that maps between joint-space (arm configuration) constraints and Cartesian-space (hand pose) denoising is challenging.

Contributions of the letter can be summarized as follows:
\begin{enumerate}[leftmargin=*]
    \item We formulate arm-aware grasp generation as a joint optimization of grasp pose and arm configuration, deriving its relation to guided sampling on the pre-trained arm-agnostic grasp diffusion model with added arm constraints.
    \item We derive analytical forms and gradients for three commonly used arm-related constraints (i.e., collision avoidance, hand reachability, and joint proximity) to construct the gradient for guidance, addressing the complex mapping between joint-space constraints and Cartesian-space denoising.
    \item We design comprehensive benchmark scenarios for simulation and real-world evaluation, featuring high obstacle coverage and grasps near arm limits, which thoroughly verify that our method generates successful grasps that satisfy constraints with a significantly higher probability than the commonly used rejection sampling strategy. The proposed approach is applicable to various robotic arms (e.g., UR5 and Franka) and environments, utilizing a single hand-centric grasp generation model.
\end{enumerate}
%

\section{Related Works}
\subsection{Dexterous Grasp Generation}
Dexterous grasp generation aims to predict feasible grasp poses for robotic hands from object meshes or point clouds, typically leveraging large-scale datasets. Early methods relied on supervised learning to directly regress grasp poses and evaluate grasp quality \cite{li2022hgc, chen2023learning}, or to construct object-centric contact representations that are later converted into grasp configurations through optimization or regression \cite{li2022contact2grasp}. To better capture the multimodal nature of feasible grasp distributions, generative models have been adopted, improving grasp diversity and generalization \cite{wei2024learning}. For instance, Li et al. \cite{li2023gendexgrasp} proposed a conditional variational autoencoder (CVAE) that jointly models hand rotation, translation, and joint articulation. More recently, diffusion models and normalizing flows have shown strong scene-conditioned distribution modeling capabilities and are rapidly becoming prevalent for grasp generation. Some recent approaches directly leverage diffusion and flow-based models to jointly generate grasp poses and their associated hand joint configurations, improving the diversity and expressiveness of grasp distributions \cite{Zhang2024DexGraspNet2L, xu2023unidexgrasp, chen2025dexonomy}. Meanwhile, other methods learn contact geometry and force distributions through diffusion, allowing the resulting grasp representations to generalize across different dexterous hand configurations \cite{lu2024ugg, ma2025contactmap}. However, despite these advances in cross-hand generalization, few methods explicitly account for arm kinematics or environmental obstacles during generation. This hand-centric assumption ultimately restricts execution success, particularly in spatially or kinematically constrained settings. While some recent works attempt to incorporate the robotic arm during data synthesis to produce arm-hand joint datasets \cite{Chen2024BODexSA}, they are typically limited to a specific arm and simple tabletop scenes, hindering generalization to different arm embodiments or diverse environments. 

\subsection{Guided Diffusion Generation in Robotics}
An advantage of diffusion models is their ability to integrate guidance during sampling, allowing the generated trajectories to satisfy task-specific constraints. Classifier guidance \cite{dhariwal2021diffusion} and classifier-free guidance (CFG) \cite{Liu2021MoreCF} steer the sampling process toward a desired category using log-probability gradients, while energy-based guidance injects the gradients of differentiable cost functions directly into the denoising steps. In robotics, guided diffusion has been successfully used in motion planning by incorporating collision penalties, kinematic constraints, and joint limits into sampling, yielding feasible trajectories in joint space \cite{luo2024potential, zhang2024robotdiffuse, saha2024edmp}. Diffusion guidance has also been applied to grasp generation. 
\blue{Weng et al. \cite{weng2024dexdiffuser} employed a learning-based evaluator for both in-sampling guidance and post-sampling refinement, whereas Zhong et al. introduced a physics-guided sampler leveraging explicit gradients of differentiable physical grasp metrics \cite{Zhong2025DexGraspAT}.}
\blue{Similarly, Ko et al. \cite{ko2025simultaneous} projected diffusion scores onto the linearized feasible region of collision constraints, though their method is restricted to parallel-jaw grippers and sparse obstacles. Beyond these physical and geometric constraints,} other works leverage language instructions as guidance to identify the target object or part, enabling semantically meaningful and functionally relevant grasping \cite{singh2024constrained, li2025language}. However, most of these methods primarily focus only on the local optimization of hand configurations to improve grasp quality, while overlooking the kinematic feasibility and environmental constraints imposed by the robotic arm bodies, especially the need for global optimization over the whole-arm configurations in highly constrained spaces.

\section{Methods}
\begin{figure*}[!tb]
    \centering
    \includegraphics[width=1\linewidth]{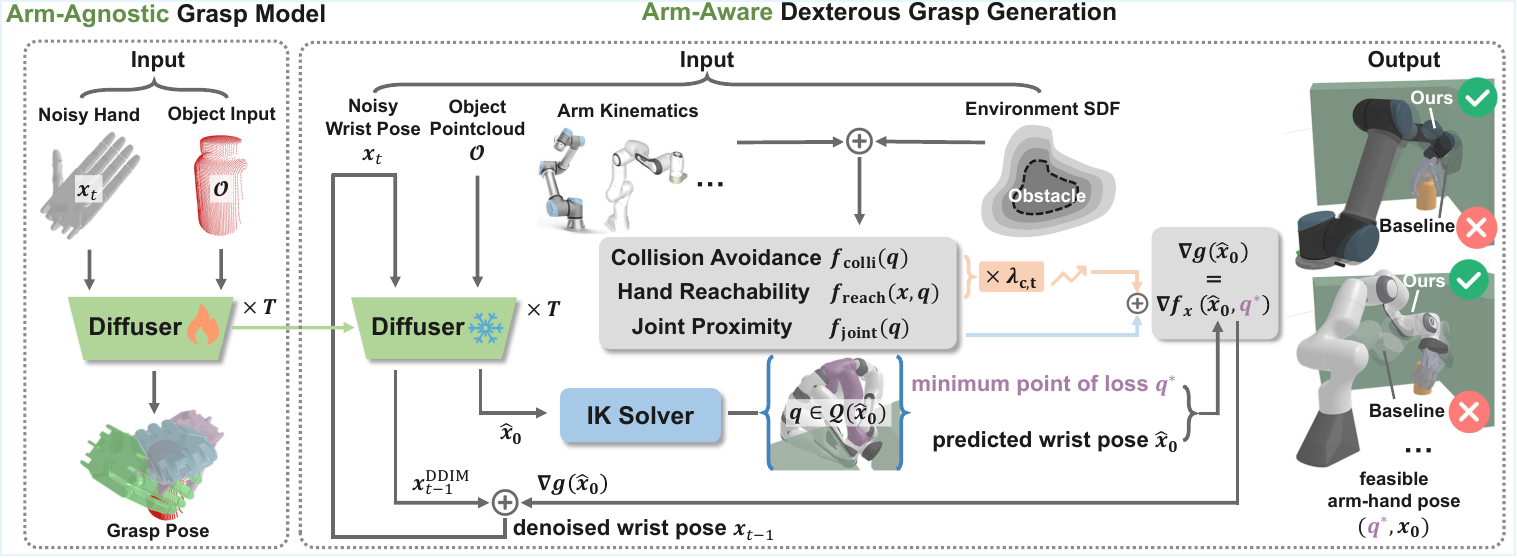}
    \caption{Overview of the proposed arm-aware dexterous grasp generation method. Initially, we pretrain an arm-agnostic diffusion model to capture the distribution of wrist poses for floating hands. During sampling, arm kinematics and environment SDF are integrated as constraints, with their gradients guiding the denoising process. This approach significantly enhances the proportion of feasible grasps, adaptable to various arm-hand configurations and constrained environments.}
    \label{fig: pipeline}
    \vspace{-15pt}
\end{figure*}
In this section, we detail the problem formulation and solution for arm-aware dexterous grasp generation, followed by the derivation of analytic forms and gradients for arm-related constraints. See Fig.~\ref{fig: pipeline} for an overview of the proposed method.
\subsection{Preliminaries}
\label{sec: method/preliminaries}
Classical dexterous grasp generation aims to learn a distribution $p(\mathcal{G} \vert \mathcal{O})$ over a synthesized grasp dataset, given the partially observed object information $\mathcal{O}$ (i.e., a single-view point-cloud) \cite{Zhong2025DexGraspAT}. A dexterous grasp $\mathcal{G}=\left(\bm{q}^{\text{hand}}, \bm{x}\right)$ contains the hand joint angles $\bm{q}^{\text{hand}} \in \mathbb{R}^{n_q}$ and the wrist pose $\bm{x} \in \mathrm{SE(3)}$. As observed in \cite{Chen2024BODexSA}, the wrist pose distribution $p(\bm{x} \vert \mathcal{O})$ typically exhibits a multi-modal pattern, which can be effectively captured by diffusion probabilistic models \cite{Ho2020DenoisingDP}. Once a wrist pose $\bm{x}$ is sampled, the joint angles $\bm{q}^{\text{hand}}$ can be deterministically predicted by a regression network. We therefore adopt a diffusion model to generate wrist poses. During training, a clean pose $\bm{x}_0$ is randomly drawn from the dataset and perturbed with Gaussian noise $\bm{\epsilon}_t$ as
\begin{equation}
    \bm{x}_t = \sqrt{\bar{\alpha}_t} \bm{x}_0 + \sqrt{1 - \bar{\alpha}_t} \bm{\epsilon}_t.
\end{equation}
Here, $\bar{\alpha}_t = \prod_{s=1}^{t} (1 - \beta_s)$ denotes the cumulative noise decay factor, and $\beta_s \in (0,1)$ determines the variance of the Gaussian noise added in diffusion step $s$. The optimization objective is derived to minimize the difference between the added noise and predicted noise
\begin{equation}
    \mathcal{L}_{\theta} := \text{MSELoss} (\bm{\epsilon}_t, \bm{\epsilon}_{\theta}(\bm{x}_t, \mathcal{O}, t)).
\end{equation}
After training, the learned noise predictor $\bm{\epsilon}_{\theta}$ is used during inference to perform iterative denoising. Specifically, we follow the DDIM sampling framework \cite{song2021denoising} to predict the clean wrist pose $\hat{\bm{x}}_0$ and update the sample $\bm{x}_{t-1}^{\text{DDIM}}$ deterministically
\begin{equation}
\hat{\bm{x}}_0 = 
\frac{1}{\sqrt{\bar{\alpha}_t}}
\left( \bm{x}_t - \sqrt{1 - \bar{\alpha}_t}\,\bm{\epsilon}_\theta(\bm{x}_t, \mathcal{O}, t) \right),
\label{x_0}
\end{equation}
\begin{equation}
\bm{x}_{t-1}^{\text{DDIM}} =
\sqrt{\bar{\alpha}_{t-1}}\, \hat{\bm{x}}_0
+ \sqrt{1 - \bar{\alpha}_{t-1}}\, \bm{\epsilon}_\theta(\bm{x}_t, \mathcal{O}, t).
\label{xt-1}
\end{equation}
The estimated $\hat{\bm{x}}_0$ is then used to enforce arm-related constraints, which will be detailed in Sec.~\ref{sec: method/problem_formulation}. From the perspective of Langevin dynamics, the DDIM update can be seen as moving the sample along the log-probability gradient \cite{luo2022understanding}
\begin{equation}
    \bm{x}_{t-1}^{\text{DDIM}} \approx \bm{x}_t + \frac{\beta_t}{2} \nabla_{\bm{x}}\log p_{\theta}(\bm{x}_t|\mathcal{O}),
    \label{eq: log_prob}
\end{equation}
i.e., toward regions of higher likelihood under the learned wrist pose distribution. After denoising, the corresponding joint angles $\bm{q}^{\text{hand}}$ are then predicted by an MLP.

\subsection{Problem Formulation of the Arm-Aware Grasp Generation}
\label{sec: method/problem_formulation}

The learned hand-centric diffusion model $p_\theta(\bm{x} \mid \mathcal{O})$ for wrist-pose generation does not ensure that a sampled pose $\bm{x} \sim p_\theta$ satisfies the constraints of the robotic arm. To address this limitation, we extend the diffusion sampling process to incorporate \textit{arm awareness}. Under arm-related constraints, some samples $\bm{x}$ lie within the feasible set, while others do not.
The key idea is that, through gradient-based guidance, \textit{nearly-feasible samples are gradually pulled toward the feasible region}, resulting in a higher proportion of feasible samples and thus forming a \textit{modified distribution}. 
Hence, our goal is to directly sample wrist poses from this \textit{modified distribution} $\pi$, subject to two requirements: 1) The modified distribution $\pi$ should remain close to the arm-agnostic distribution $p_\theta(\bm{x} \mid \mathcal{O})$ to maintain grasp quality; 2) Each wrist pose $\bm{x}$ must be reachable by the robotic arm while satisfying arm-related constraints.

We formulate this as a bi-level optimization problem that separately optimizes the diffusion sampling distribution $\pi$ and the arm configuration $\bm{q}$. The outer optimization aims to update the diffusion sampling distribution $\pi$ by minimizing the Kullback–Leibler (KL) divergence to the pretrained model $p_{\theta}$, together with the expected arm-aware loss $g(\bm{x})$ computed at each wrist pose $\bm{x} \sim \pi$. Formally, the outer problem is defined as a variational optimization over $\pi$
\begin{equation}
\begin{aligned}
    \min_{\pi} \mathrm{KL} (\pi \Vert p_\theta) + \lambda_g\mathbb{E}_{x \sim \pi} \big [g(\bm{x}) \big],
    \end{aligned}
    \label{eq: KL_based_constrained_sampling}
\end{equation}
where the value function $g(\bm{x})$ represents the minimum loss achievable by the robotic arm under feasibility constraints and $\lambda_g$ is a positive weight coefficient. For each sampled wrist pose $\bm{x}$, we first consider the following inner constrained optimization over the joint configuration $\bm{q}$
\begin{equation}
    g(\bm{x}) = \min_{q \in \mathcal{Q}(\bm{x})} f_s(\bm{x}, \bm{q}) \quad \text{s.t.} \quad f_c(\bm{x}, \bm{q}) \leq 0,
    \label{eq: inner_optimization}
\end{equation}
where 
\begin{equation}
    \mathcal{Q}(\bm{x}) = \{\bm{q} \;\big|\; \mathrm{FK}(\bm{q}) = \bm{x}\}
    \label{eq: ik_solution_set}
\end{equation}
denotes the feasible configuration set defined by forward kinematics $\mathrm{FK}(\cdot)$.
\blue{The terms $f_s$ and $f_c$ correspond to the soft penalty and hard inequality constraint, respectively. In this letter, collision avoidance and hand reachability are enforced as hard constraints, whereas joint proximity is modeled as a soft penalty (see Sec.~\ref{sec: method/cartesian_space_cost})}.
For the inner optimization (\ref{eq: inner_optimization}), we introduce the Lagrangian to handle the inequality constraint
\begin{equation}
\begin{aligned}
\mathcal{L}(\bm{x}, \bm{q}; \lambda_c) =  f_s(\bm{x}, \bm{q}) + \lambda_c f_c(\bm{x}, \bm{q}),
\end{aligned}
\label{eq:lagrangian}
\end{equation} 
where $\lambda_c \ge 0$ is the Lagrange multiplier associated with the inequality constraint. 
The optimal arm configuration is then obtained by minimizing this Lagrangian within the feasible set
\begin{equation}
\begin{aligned}
\bm{q}^{*}
& =\arg\min_{\bm{q}\in \mathcal{Q}(\bm{x})} f(\bm{x}, \bm{q}), \\
\text{where} \:\: & f(\bm{x}, \bm{q})  = f_s(\bm{x}, \bm{q})
+\lambda_c f_c(\bm{x}, \bm{q}).
\end{aligned}
\label{eq:inner_opt}
\end{equation}
Following the primal–dual method \cite{boyd2004convex}, the dual variable for the inequality constraint is updated via gradient ascent on the Lagrangian in (\ref{eq:lagrangian})
\begin{equation}
\lambda_{c,{t-1}} = \lambda_{c,t} + \eta_{\lambda} f_c(\bm{x}, \bm{q}^*),
\label{eq: weight updated}
\end{equation}
where $\eta_\lambda > 0$ is the learning rate. This update gradually increases the penalty on infeasible wrist poses as denoising proceeds, thereby enforcing the hard constraint more strictly.
This inner optimization thus defines the value function $g(\bm{x}) = f(\bm{x}, \bm{q}^*)$, which then guides the outer update of the diffusion distribution over wrist poses. 
Following \cite{Zhang2025ConstrainedDF}, solving the variational optimization in (\ref{eq: KL_based_constrained_sampling}) yields a pointwise closed-form solution
\begin{equation}
\pi^*(\bm{x})\:\propto\:
p_\theta(\bm{x} | \mathcal{O})\,
\exp\!\big[ -g(\bm{x})
\big],
\label{eq:tilted_distribution}
\end{equation}
which corresponds to an exponentially tilted version of the pretrained distribution. By descending the negative log-probability of this modified distribution in (\ref{eq:tilted_distribution}) with the substitution of DDIM update (\ref{eq: log_prob}), we obtain the following denoising update
\begin{equation}
\bm{x}_{t-1} \;=\; \bm{x}_{t-1}^{\mathrm{DDIM}} 
- \frac{\beta_t}{2}\Big[ \nabla g(\hat{\bm{x}}_0)\Big],
\label{eq: primal_update_constrained_sampling}
\end{equation}
where $\bm{x}_{t-1}^{\text{DDIM}}$ is obtained from (\ref{xt-1}). This update resembles guided sampling, integrating both the original denoising process and guidance from arm-related feasibility constraints.
Compared to \cite{Zhang2025ConstrainedDF}, we compute gradients using the predicted wrist pose $\hat{\bm{x}}_0$ instead of the noised sample $\bm{x}_t$, ensuring smoother and more consistent gradient updates \footnote{Compared to $\nabla g(\bm{x})$, $\nabla g(\hat{\bm{x}}_0)$ changes more gently, especially when considering the collision avoidance constraints.}.

\subsection{Analytical Formulation of Arm-Related Constraints}
\label{sec: method/cartesian_space_cost}
Building on the hierarchical formulation in Sec.~\ref{sec: method/problem_formulation}, we now describe how the arm-related constraints are incorporated into each denoising step (\ref{eq: primal_update_constrained_sampling}). The feasible set $\mathcal{Q}(\bm{x})$ is obtained via IK mapping from wrist poses to arm configurations. To handle multiple IK branches efficiently, we use the geometry-based analytical solver \cite{Elias2022IKGeoUR}. After obtaining the optimal joint configuration $\bm{q}^{*}$ over the feasible set according to (\ref{eq:inner_opt}), the gradient of the value function can then be computed as  

\begin{equation}
\nabla g(\bm{x}) = \nabla_{\bm{x}} f\left(\bm{x}, \bm{q}^*\right).
\label{eq: gradient_g}
\end{equation}
We further evaluate this gradient by applying the chain rule
\begin{equation}
\nabla g(\bm{x}) 
= {\frac{\partial f}{\partial \bm{x}}(\bm{x}, \bm{q}^{*})}
+ {\left(\frac{d \bm{q}^{ *}}{d \bm{x}}\right)^{\!\top} 
\frac{\partial f}{\partial \bm{q}}(\bm{x}, \bm{q}^{ *})}.
\label{eq: chain}
\end{equation}
To compute the second term, observe that the IK solution $\bm{q}^{*}$ can be viewed as the local minimizer of the following quadratic program, with $\bm{q}^{\text{ref}}$ denoting a reference configuration
\begin{equation}
\bm{q}^{*} 
= \arg\min_{\bm{q}} 
\frac{1}{2}\| \text{FK}(\bm{q}) - \bm{x} \|_2^2 
+ \frac{\lambda_q^2}{2}\|\bm{q} - \bm{q}_{\mathrm{ref}}\|_2^2 .
\label{eq: least_square_of_IK}
\end{equation}
Here, $\lambda_q$ serves as a damping coefficient that stabilizes the solution.
Through sensitivity analysis of (\ref{eq: least_square_of_IK}), the gradient of $\bm{q}^{*}$ can be derived as the damped pseudo-inverse, which is independent of the reference configuration $\bm{q}^{\text{ref}}$
\begin{equation}
\frac{d \bm{q}^{*}}{d \bm{x}} 
= \bm{J}_{\lambda_q}^{\dagger} 
= (\bm{J}^{\top} \bm{J} + \lambda_q^2 \bm{I})^{-1} \bm{J}^{\top},
\end{equation}
where $\bm{J}$ denotes the geometric Jacobian of the arm. Substituting this result back into (\ref{eq: chain}), the closed-form gradient with respect to the wrist pose is obtained as
\begin{equation}
\nabla g(\bm{x}) 
= {\frac{\partial f}{\partial \bm{x}}(\bm{x}, \bm{q}^{ *})}
+ { (\bm{J}_{\lambda_q}^{\dagger})^{\top}
\frac{\partial f}{\partial \bm{q}}(\bm{x}, \bm{q}^{ *})}.
\label{eq: grad}
\end{equation}
This indicates that the arm-related gradient comprises a direct term with respect to the wrist pose $\bm{x}$ and an indirect term propagated through the joint configuration $\bm{q}$ via the damped Jacobian pseudo-inverse.

Based on this gradient formulation, we now present the specific forms of the three arm-related constraints.

\textbf{Hand Reachability}: This constraint ensures that the wrist pose remains within the reachable workspace of the arm
by penalizing the forward kinematics error
\begin{equation}
    f_{\text{reach}}(\bm{x}, \bm{q}) = \frac{1}{2}\|\text{FK}(\bm{q}) - \bm{x} \|^2.
\end{equation}
The partial derivatives of the reachability loss are written as
\begin{equation}
 \frac{\partial f_{\text{reach}}}{\partial \bm{x}} = \bm{x}-\mathrm{FK}(\bm{q}), \; \frac{\partial f_{\text{reach}}}{\partial \bm{q}} = \bm{J}^{\top}(\mathrm{FK}(\bm{q})-\bm{x}).
\label{eq: cartesian_grad_reach}
\end{equation}

\textbf{Arm–Environment Collision Avoidance}:  This constraint prevents potential collisions between the robotic arm and the environment. It is determined by the minimum value $d(\bm{q})$ of the points on the arm’s sphere-based collision model within the environment’s signed distance field (SDF). \blue{To obtain smooth collision gradients, we adopt a SoftPlus activation function whose derivative yields a sigmoid-shaped distance-dependent scaling. Specifically, the collision cost is defined as
\begin{equation}
f_{\text{colli}}(\bm{q})
=  \frac{1}{\kappa}\log\Big(1 + e^{\kappa d}\Big) - d,
\end{equation}
where $\kappa$ controls the sharpness of the transition. Taking the gradient with respect to $\bm{q}$ gives
\begin{equation}
    \frac{\partial f_{\text{colli}}}{\partial \bm{q}} = \left[\sigma(\kappa d)-1\right] \bm{J}_p^{\top}\bm{n},
    \label{eq: cartesian_grad_colli}
\end{equation}
where $\sigma(z) = \frac{1}{1 + e^{-z}}$ denotes the sigmoid function, $\bm{J}_p \in \mathbb{R}^{3 \times n}$ is the translational Jacobian of the point on the robotic arm with the minimum SDF value, and $\bm{n}$ is the unit vector which projects $\bm{J}_p$ onto the SDF gradient.}

\textbf{Joint Proximity}:
To reduce unnecessary joint-space movement, we introduce a penalty on the deviation of the solved arm configuration $\bm{q}$ from the current configuration $\bm{q}_{\text{cur}}$
\blue{
\begin{equation}
    f_{\text{joint}}(\bm{q}) = \frac{1}{2}\|\bm{q} - \bm{q}_{\text{cur}}\|^2,
\end{equation}
The corresponding gradient is given by 
\begin{equation}
    \frac{\partial f_{\text{joint}}}{\partial \bm{q}} 
    = \bm{q} - \bm{q}_{\text{cur}}.
    \label{eq: cartesian_grad_joint}
\end{equation}}

\blue{
The \textbf{Hand Reachability} and \textbf{Collision Avoidance} terms are treated as hard feasibility constraints, whereas the \textbf{Joint Proximity} is considered a soft penalty in (\ref{eq: inner_optimization}). Thus, we have
\begin{equation}
    f(\bm{x},\bm{q})=\lambda_{c}\left[\alpha f_{\text{reach}}(\bm{x},\bm{q})+(1-\alpha)f_{\text{colli}}(\bm{q})\right]+ \lambda_sf_{\text{joint}}(\bm{q}).
    \label{eq: f_x_q_specific_form}
\end{equation}
where $\alpha \in (0,1)$ balances the two hard constraints, and $\lambda_s > 0$ controls the strength of the joint proximity penalty. 
As described in Sec.~\ref{sec: method/problem_formulation}, the constraint weight $\lambda_c$ is dynamically updated with learning rate $\eta_\lambda$ following (\ref{eq: weight updated}). A sensitivity analysis of $\lambda_c$, $\eta_\lambda$, $\lambda_s$, and other key hyper-parameters is provided in Appendix~A-B.}
Finally, the gradient $\nabla g(\bm{x})$ required by (\ref{eq: primal_update_constrained_sampling}) can be computed with (\ref{eq: grad}), (\ref{eq: cartesian_grad_reach}), (\ref{eq: cartesian_grad_colli}) and (\ref{eq: cartesian_grad_joint}).
The overall grasp generation procedure is summarized in Alg.~\ref{algo: arm_aware_dex_grasp_generation}.
Besides, the roles of different constraints are further validated through the experiments in Sec.~\ref{sec: experiments}.

\begin{algorithm}[t]
\caption{Arm-Aware Dexterous Grasp Generation}
\SetAlgoLined
\KwIn{noise predictor $\epsilon_\theta$ of the pre-trained diffusion grasp generation model, environment SDF, object partial observation $\mathcal{O}$, initial value and learning rate of $\lambda_c$, current configuration $\bm{q}_{\text{cur}}$}

\For{$t$ \textnormal{from} $T_N$ \KwTo $1$} {
Compute $\bm{x}_{t-1}^{\text{DDIM}}$ using the denoising step (\ref{xt-1})

Compute the IK solution set $\mathcal{Q}$ (\ref{eq: ik_solution_set}) and the cost function $f(\bm{x}, \bm{q})$ according to (\ref{eq: f_x_q_specific_form})

Compute $\bm{q}^{*}$ according to (\ref{eq:inner_opt})

Update $\lambda_{c}$ with (\ref{eq: weight updated})


Compute $\nabla g(\bm{x})$ with  (\ref{eq: grad}), (\ref{eq: cartesian_grad_reach}), (\ref{eq: cartesian_grad_colli}), and (\ref{eq: cartesian_grad_joint})

Update the noised sample $\bm{x}_{t-1}$ with (\ref{eq: primal_update_constrained_sampling})
}
\Return{$\bm{x}_0, \bm{q}^{*}$}
\label{algo: arm_aware_dex_grasp_generation}
\end{algorithm}

\section{Experiments}
\label{sec: experiments}
We design the experiments to answer the following questions: 1) 
Can the proposed method generate feasible grasps that satisfy arm-related constraints with a higher probability?
2) How does incorporating these constraints affect physical grasp quality? 3) Is the proposed method easily adaptable to different robotic arms and environments without additional data synthesis or training? \blue{Additional evaluation results can be found in the Appendix (available on our \href{https://arm-aware-dexgrasp.github.io/}{Project Website}).} 

\textbf{Experiment Setup.} We use the Shadow Hand in simulations and the LEAP Hand for real-world evaluations, considering two common robotic arms, UR5 and Franka. We assume the environment can be represented as a combination of closed geometries, enabling a well-defined SDF. We adopt the sphere robot collision models from cuRobo \cite{Sundaralingam2023CuRoboPC} for collision avoidance. We train the hand-centric diffusion model $p_\theta$ on roughly 50k tabletop grasps synthesized with BODex \cite{Chen2024BODexSA} on the processed DGN dataset \cite{chen2025dexonomy}. The simulation evaluation is performed using DGN's test set, containing 10,892 objects with varying geometries, poses and scales. For each object, 10 random grasp poses and arm configurations are sampled, yielding a total of approximately 100k grasps.

\textbf{Evaluation Metrics.} \blue{For evaluation, each sampled grasp pose uses the IK solution with the lowest constraint loss $f(\bm{x},\bm{q})$ defined as (\ref{eq:inner_opt}). This criteria is kept the same across all methods, separating the influence of IK selection on the evaluation results.} We consider four types of metrics: 
1) \textit{Constraint Satisfaction}. For the hard constraints, we filter out grasps that violate these constraints and report the percentage of arm-feasible grasps among all sampled grasps that satisfy the Collision Avoidance constraint (Collision Feasible Rate, CFR), the Hand Reachability constraint (Reachability Feasible Rate, RFR), and both (Feasible Rate, FR). For the soft constraint, we report the average Joint Proximity loss of all successful grasps (Average Joint Proximity, AJP). 
2) \textit{Grasp Success Rate} (GSR). Among all feasible grasps that satisfy the hard constraints after filtering, we report the percentage of grasps that are successfully executed in MuJoCo. This metric indicates the quality of the arm-feasible grasps.
3) \textit{Success Rate} (SR). This metric, akin to GSR, measures the percentage of feasible and successful grasps from all sampled grasps and is equal to the product of FR and GSR.
4) \textit{Object Success Rate} (OSR). We report the percentage of the 10,892 objects that, among all sampled grasps, have at least one that is both feasible and successfully executed in MuJoCo. 

\textbf{Baselines.}
In the experiments, the \textit{Baseline (RS)} refers to rejection sampling (RS) from the pretrained grasp distribution $p_\theta$.
\blue{Moreover, we compare our approach against two representative constraint-aware diffusion-based methods (Sec.~\ref{subsec: comparison_other_constraint_aware_generation}): 1) \textit{Projected Denoising (PD)} \cite{ko2025simultaneous}, which projects denoising directions onto the linearized feasible region via optimization-based differential IK; and 2) \textit{Post-Sampling Refinement (PSR)} \cite{weng2024dexdiffuser}, which employs constraint gradients to locally refine the sampled grasps. To unify differing baseline setups, we re-implemented them using our diffusion backbone, analytical gradients, and evaluation protocol for a more rigorous comparison.}
\vspace{-5pt}
\subsection{Simulation Studies}
\label{sec: simulation_studies}

\begin{table}[!tb]
\centering
\caption{Performance of grasp generation in constrained environments (UR5)}
\resizebox{\linewidth}{!}{
\begin{threeparttable}
    \begin{tabular}{ccccc@{\hspace{7pt}}c@{\hspace{3pt}}c@{\hspace{3pt}}c}
    \toprule
                                     &                                       & \multicolumn{3}{c}{\textbf{C.   Satisfaction(\%)↑}}                        &          \\ \cmidrule(lr{0pt}){3-5}  
    \multirow{-2.5}{*}{\textbf{Scene}} & \multirow{-2.5}{*}{\textbf{Method}}     & CFR                      &  RFR                      & FR                       & \multirow{-2.5}{*}{\begin{tabular}{c} GSR \\ (\%)↑ \end{tabular}}                      &
    \multirow{-2.5}{*}{\begin{tabular}{c} SR \\ (\%)↑ \end{tabular}}    &
    \multirow{-2.5}{*}{\begin{tabular}{c} OSR \\ (\%)↑ \end{tabular}}                      \\ \midrule
                                     & Baseline (RS)  & 15.31 & 99.96 & 15.31 &  \textbf{58.17} & 8.91 & 55.15 \\
    \multirow{-2}{*}{S1}             & \cellcolor[HTML]{D9D9D9}\textbf{Ours} & \cellcolor[HTML]{D9D9D9}\textbf{88.87} & \cellcolor[HTML]{D9D9D9}\textbf{99.99} & \cellcolor[HTML]{D9D9D9}\textbf{88.87} &  \cellcolor[HTML]{D9D9D9}53.08 & \cellcolor[HTML]{D9D9D9}\textbf{47.17} & \cellcolor[HTML]{D9D9D9}\textbf{91.77} \\ \midrule
                                     & Baseline (RS)  & 31.08 & 99.24 & 31.08 &  \textbf{60.08} & 18.68 & 74.32 \\
    \multirow{-2}{*}{S2}             & \cellcolor[HTML]{D9D9D9}\textbf{Ours} & \cellcolor[HTML]{D9D9D9}\textbf{87.91} & \cellcolor[HTML]{D9D9D9}\textbf{99.95} & \cellcolor[HTML]{D9D9D9}\textbf{87.91} &  \cellcolor[HTML]{D9D9D9}49.11 & \cellcolor[HTML]{D9D9D9}\textbf{43.17} & \cellcolor[HTML]{D9D9D9}\textbf{91.54} \\ \midrule
                                     & Baseline (RS)  & 25.75 & 72.24 & 25.75 &  \textbf{58.46} & 15.05 & 70.31 \\
    \multirow{-2}{*}{S3}             & \cellcolor[HTML]{D9D9D9}\textbf{Ours} & \cellcolor[HTML]{D9D9D9}\textbf{62.89} & \cellcolor[HTML]{D9D9D9}\textbf{99.13} & \cellcolor[HTML]{D9D9D9}\textbf{62.89} & \cellcolor[HTML]{D9D9D9}51.14 & \cellcolor[HTML]{D9D9D9}\textbf{32.16} & \cellcolor[HTML]{D9D9D9}\textbf{89.11} \\ \bottomrule
    \end{tabular}
\end{threeparttable}}
\label{tab: collision_graspgen_result_ur5}
\end{table}

\begin{table}[!tb]
\centering
\caption{Performance of grasp generation in constrained environments (Franka)}
\resizebox{\linewidth}{!}{
    \begin{tabular}{ccccc@{\hspace{7pt}}c@{\hspace{3pt}}c@{\hspace{3pt}}c}
    \toprule
                                     &                                       & \multicolumn{3}{c}{\textbf{C. Satisfaction(\%)↑}}                        &          \\ \cmidrule(lr{0pt}){3-5}  
    \multirow{-2.5}{*}{\textbf{Scene}} & \multirow{-2.5}{*}{\textbf{Method}}     & CFR                      &  RFR                      & FR                       & \: \multirow{-2.5}{*}{\begin{tabular}{c} GSR \\ (\%)↑ \end{tabular}}                      & \multirow{-2.5}{*}{\begin{tabular}{c} SR \\ (\%)↑ \end{tabular}} & \multirow{-2.5}{*}{\begin{tabular}{c} OSR \\ (\%)↑ \end{tabular}}                      \\ \midrule
                                     & Baseline (RS)  & 10.92 &  85.22 & 10.58 & \textbf{54.18} & 5.73 & 39.07 \\
    \multirow{-2}{*}{S1}             & \cellcolor[HTML]{D9D9D9}\textbf{Ours} & \cellcolor[HTML]{D9D9D9}\textbf{61.03} & \cellcolor[HTML]{D9D9D9}\textbf{85.95} & \cellcolor[HTML]{D9D9D9}\textbf{59.18} & \cellcolor[HTML]{D9D9D9}48.50 & \cellcolor[HTML]{D9D9D9}\textbf{28.70} & \cellcolor[HTML]{D9D9D9}\textbf{84.76} \\ \midrule
                                     & Baseline (RS)  & 32.04 &  85.35 & 30.31 & \textbf{57.82} & 17.53 & 73.60 \\
    \multirow{-2}{*}{S2}             & \cellcolor[HTML]{D9D9D9}\textbf{Ours} & \cellcolor[HTML]{D9D9D9}\textbf{85.87} &  \cellcolor[HTML]{D9D9D9}\textbf{91.86} & \cellcolor[HTML]{D9D9D9}\textbf{83.28} & \cellcolor[HTML]{D9D9D9}47.32 & \cellcolor[HTML]{D9D9D9}\textbf{39.41} & \cellcolor[HTML]{D9D9D9}\textbf{90.96} \\ \midrule
                                     & Baseline (RS)  & 32.56 &  55.69 & 32.37 & \textbf{57.09} & 18.48 & 75.47 \\
    \multirow{-2}{*}{S3}             & \cellcolor[HTML]{D9D9D9}\textbf{Ours} & \cellcolor[HTML]{D9D9D9}\textbf{72.29} &  \cellcolor[HTML]{D9D9D9}\textbf{84.84} & \cellcolor[HTML]{D9D9D9}\textbf{72.01} & \cellcolor[HTML]{D9D9D9}49.19 & \cellcolor[HTML]{D9D9D9}\textbf{35.42} & \cellcolor[HTML]{D9D9D9}\textbf{89.64} \\
    \midrule
                                     & Baseline (RS)  & 20.01 &  90.64 & 19.91 & \textbf{53.99} & 10.75 & 56.45 \\
    \multirow{-2}{*}{S4}             & \cellcolor[HTML]{D9D9D9}\textbf{Ours} & \cellcolor[HTML]{D9D9D9}\textbf{59.02} & \cellcolor[HTML]{D9D9D9}\textbf{96.50} & \cellcolor[HTML]{D9D9D9}\textbf{58.73} & \cellcolor[HTML]{D9D9D9}44.86 & \cellcolor[HTML]{D9D9D9}\textbf{26.35} & \cellcolor[HTML]{D9D9D9}\textbf{83.45} \\\bottomrule
    \end{tabular}
    }
    \label{tab: collision_graspgen_result_franka}
    \vspace{-15pt}
\end{table}

It is worth noting that Collision Avoidance is essential for constrained grasping, while Hand Reachability and Joint Proximity play critical roles for their respective purposes. We first evaluate Collision Avoidance in Sec.~\ref{sec: sim_collision_avoidance}, followed by an assessment of the other two constraints and their interaction with Collision Avoidance in Secs.~\ref{sec: sim_hand_reachability} and \ref{sec: sim_joint_proximity}.

\subsubsection{Performance of the Collision Avoidance Constraint}
\label{sec: sim_collision_avoidance}
To evaluate the performance of guided grasp generation in constrained environments, we design four challenging scenes (S1$\sim$S4) with high obstacle coverage, as shown in Appendix~A-A. The feasible regions in these scenarios are relatively small, which significantly impacts the efficiency of rejection sampling. The results are shown in Tab.~\ref{tab: collision_graspgen_result_ur5} and Tab.~\ref{tab: collision_graspgen_result_franka}. On the one hand, the proposed method significantly enhances constraint satisfaction, particularly with regard to the CFR and FR.
On the other hand, the incorporation of guidance slightly degrades quality of the feasible grasps, as indicated by GSR, by altering the direction in the original denoising process. 
It is worth noting that, while grasp quality may be affected, the proposed method significantly improves SR and the probability of achieving at least one successful grasp for a given object—measured by the Object Success Rate (OSR)—by increasing the proportion of feasible grasps. In Appendix~A-C, we explore a preliminary solution for maintaining grasp quality through null-space projection, which merits further investigation.
Additionally, the proposed method demonstrates consistent performance on both UR5 (Tab.~\ref{tab: collision_graspgen_result_ur5}) and Franka (Tab.~\ref{tab: collision_graspgen_result_franka}), demonstrating its adaptability to robotic arms with different kinematics.
For Franka, both the proposed method and the baseline show more noticeable constraint violations on S1 (Tab.~\ref{tab: collision_graspgen_result_franka}), likely due to Franka's larger collision volume and limited workspace, which reduce the feasible region.
%

%
%

\begin{table}[!b]
\vspace{-10pt}
\centering
\caption{Performance of reachable grasp generation near workspace boundaries}
\resizebox{\linewidth}{!}{
\begin{threeparttable}
    \begin{tabular}{c@{\hspace{8pt}}cccc@{\hspace{6pt}}c@{\hspace{2pt}}c@{\hspace{2pt}}c}
    \toprule
                                     &                                       & \multicolumn{3}{c}{\textbf{C. Satisfaction(\%)↑}}                        &          \\ \cmidrule(l){3-5} 
    \multirow{-2.5}{*}{\textbf{Arm}} & \multirow{-2.5}{*}{\textbf{Method}}     & CFR                      &  RFR                      & FR                       & \multirow{-2.5}{*}{\begin{tabular}{c} GSR \\ (\%)↑ \end{tabular}}                      & \multirow{-2.5}{*}{\begin{tabular}{c} SR \\ (\%)↑ \end{tabular}}                      & \multirow{-2.5}{*}{\begin{tabular}{c} OSR \\ (\%)↑ \end{tabular}}                      \\ \midrule
                 & Baseline (RS)  & 90.06 &  26.07 & 25.74 & \textbf{59.54} & 15.32 & 66.33 \\
                 & \cellcolor[HTML]{D9D9D9}\textbf{Ours} & \cellcolor[HTML]{D9D9D9}99.25 & \cellcolor[HTML]{D9D9D9}65.58 & \cellcolor[HTML]{D9D9D9}65.57 & \cellcolor[HTML]{D9D9D9}53.04 & \cellcolor[HTML]{D9D9D9}\textbf{34.79} & \cellcolor[HTML]{D9D9D9}\textbf{89.42} \\ 
    \multirow{-2}{*}{UR5}
                 & w/o Reachability\tnote{*}   & \textbf{99.83} & 25.89 & 22.49 & 59.48 & 15.39 & 66.92 \\
                & w/o Collision\tnote{*}  & 95.77 &  \textbf{67.30} & \textbf{66.13} & 52.61 & 34.79 & 89.41 \\\midrule
                                    
                & Baseline (RS)  & 78.20 &  21.44  & 12.26 & \textbf{58.55} & 7.18 & 37.97 \\
                & \cellcolor[HTML]{D9D9D9}\textbf{Ours} & \cellcolor[HTML]{D9D9D9}96.95 &  \cellcolor[HTML]{D9D9D9}42.44 & \cellcolor[HTML]{D9D9D9}\textbf{42.08} & \cellcolor[HTML]{D9D9D9}51.06 & \cellcolor[HTML]{D9D9D9}\textbf{21.48} & \cellcolor[HTML]{D9D9D9}\textbf{76.48}  \\
                \multirow{-2}{*}{Franka}
                & w/o Reachability\tnote{*} & \textbf{99.7} &  20.16 & 20.06 & 54.15 & 10.86 & 52.72 \\
                & w/o Collision\tnote{*}  & 64.67 & \textbf{52.99}  & 30.72 & 54.01 & 16.59 & 61.39
   \\\bottomrule
    \end{tabular}
\begin{tablenotes}
\footnotesize
    \item[*] \blue{The 'w/o' variants represent ablations that omit the guidance of the associated constraint from our method.}
\end{tablenotes}
\end{threeparttable}
}
\label{tab: ik_graspgen_results}
\vspace{-10pt}
\end{table}

\begin{figure*}[!tb]
\vspace{-10pt}
    \centering
    \includegraphics[width=1\linewidth]{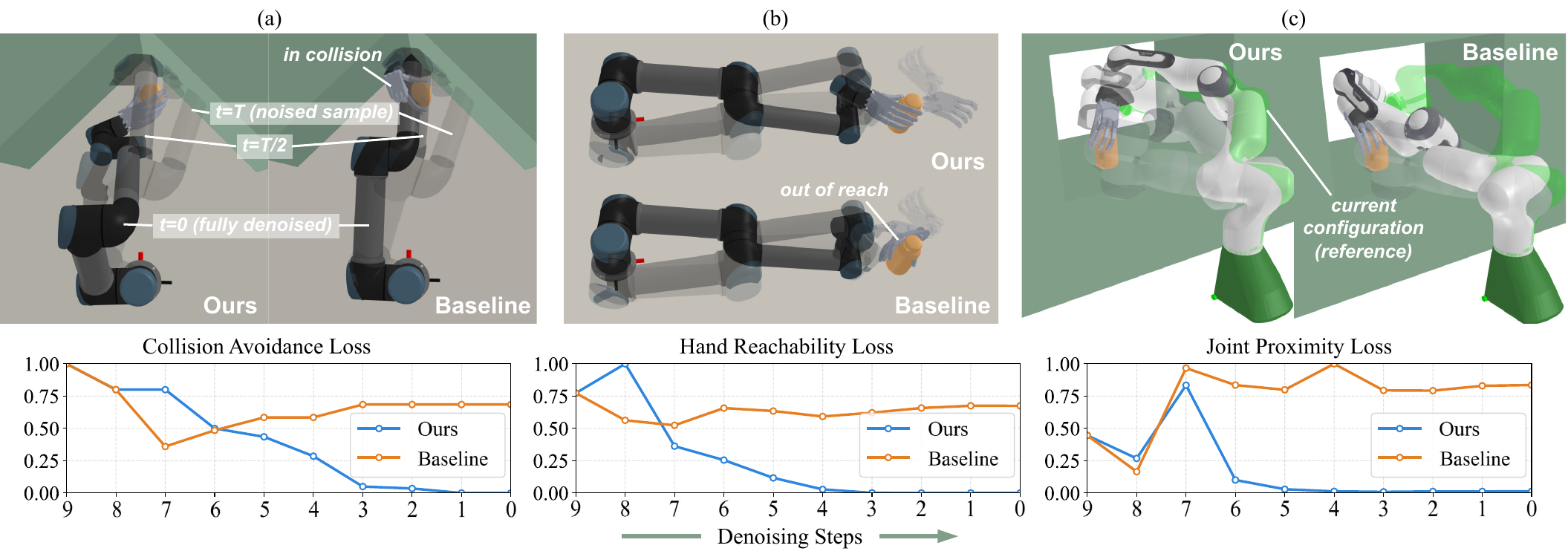}
    \vspace{-10pt}
    \caption{Illustration of representative cases highlighting the effects of various constraints. The top row illustrates the arm configurations $\bm{q}^{*}$ during the denoising process of a certain noised sample both with our method and Baseline (RS), where $\bm{q}^{*}$ at $t = T, T/2, 0$ are visualized from transparent to opaque. The denoising process 
    starts from the same initial sample shown with the highest transparency. The fully denoised hand and arm configuration are depicted with the lowest transparency.
    For clarity of illustration, we only display the evolution of hand configuration in (b).
    The green arm in (c) represents the reference joint configuration $\bm{q}_{\text{cur}}$ based on a hypothetical previous grasp attempt. The bottom row presents the evolution of normalized constraint losses over time.}
    \label{fig: case_study_all}
    \vspace{-10pt}
\end{figure*}

\subsubsection{Performance of the Hand Reachability Constraint}
\label{sec: sim_hand_reachability}
To evaluate our method's ability to generate grasps with higher reachability, we randomly place objects near the edge of the robotic arm’s workspace, as illustrated in the tabletop scene in Fig.~\ref{fig: case_study_all} (b). Under these placements, the arm can only perform grasps aligned with the object’s front-facing direction and cannot reach grasp poses from behind, while still needing to avoid collisions with the tabletop. As shown in Tab.~\ref{tab: ik_graspgen_results}, applying only the Collision Avoidance constraint (w/o Reachability) or only the Hand Reachability constraint (w/o Collision) results in optimal performance for their respective metrics (CFR and RFR). However, this comes at the expense of the other metrics, indicating that a solution addressing both constraints more effectively remains to be found. In contrast, our method incorporates both constraints during inference, guiding sampling toward feasible regions that satisfy both constraints. This leads to a substantial increase in FR and further improves OSR.

\subsubsection{Performance of the Joint Proximity Constraint}
\label{sec: sim_joint_proximity}
To evaluate our method's ability to generate grasps with arm configuration proximity, we randomly place objects on a shelf and define a reference joint configuration based on a hypothetical previous grasp attempt, as visualized in Fig.~\ref{fig: case_study_all} (c). The results in Tab.~\ref{tab: joint_proximity_results} indicate that our method reduces joint proximity loss (AJP) and delivers the best performance in both CFR and FR. We observe that applying only the Joint Proximity constraint (w/o Collision) also improves collision avoidance performance (CFR), as the reference 
configuration is collision-free and implicitly guides sampling toward safer regions.
We further demonstrate potential applications of this constraint through real-world experiments in Appendix~B.

\begin{table}[!b]
\vspace{-10pt}
\centering
\caption{Performance of grasp generation with reference joint configuration}
\resizebox{\linewidth}{!}{
\begin{threeparttable}
    \begin{tabular}{c@{\hspace{8pt}}c@{\hspace{4pt}}c@{\hspace{4pt}}c@{\hspace{0.5pt}}c@{\hspace{5pt}}c@{\hspace{2pt}}c@{\hspace{2pt}}c}
    \toprule
    \multirow{3}{*}{\textbf{Arm}} & \multirow{3}{*}{\textbf{Method}} & \multicolumn{3}{c}{\textbf{C. Satisfaction}} & \multirow{3}{*}{\begin{tabular}{c}GSR\\ (\%)↑\end{tabular}} & \multirow{3}{*}{\begin{tabular}{c}SR\\ (\%)↑\end{tabular}} & \multirow{3}{*}{\begin{tabular}{c}OSR\\ (\%)↑\end{tabular}} \\ \cline{3-5}
                     &                         & \multirow{2.3}{*}{\begin{tabular}{c} CFR \\ (\%)↑ \end{tabular}}        & \multirow{2.3}{*}{\begin{tabular}{c} FR \\ (\%)↑ \end{tabular}}        & \multirow{2.3}{*}{\begin{tabular}{c} AJP \\ ($\text{rad}^2$)↓ \end{tabular}}        &                                                                     &                                                                    &                                                                     \\
                     &                         &      &     &     &                                                                     &                                                                    &     \\                                                               
    \midrule
                                     & Baseline (RS)  & 35.55 & 34.28 & 0.0686 & \textbf{58.51} & 20.06 &71.63 \\
                 & \cellcolor[HTML]{D9D9D9}\textbf{Ours} & \cellcolor[HTML]{D9D9D9}\textbf{76.45} & \cellcolor[HTML]{D9D9D9}\textbf{74.76} & \cellcolor[HTML]{D9D9D9}0.0398 & \cellcolor[HTML]{D9D9D9}49.79 & \cellcolor[HTML]{D9D9D9} \textbf{37.72} & \cellcolor[HTML]{D9D9D9} \textbf{90.26} \\ 
    \multirow{-2}{*}{UR5}
                & w/o Collision\tnote{*}  & 45.11 & 48.83 & \textbf{0.0372} & 52.07 & 23.34 & 77.25 \\
                & w/o Joint\tnote{*}  & 49.34 & 46.49 & 0.0614 & 56.95 & 26.48 & 83.92 \\\midrule
                                    
                & Baseline (RS)  & 45.52 & 44.41 & 0.0778 & \textbf{55.08} & 24.46 & 65.38 \\

                 & \cellcolor[HTML]{D9D9D9}\textbf{Ours} & \cellcolor[HTML]{D9D9D9}\textbf{87.72} & \cellcolor[HTML]{D9D9D9}\textbf{85.12} & \cellcolor[HTML]{D9D9D9}\textbf{0.0643} & \cellcolor[HTML]{D9D9D9}42.38 & \cellcolor[HTML]{D9D9D9}\textbf{36.07} & \cellcolor[HTML]{D9D9D9}\textbf{82.93}  \\
                \multirow{-2}{*}{Franka}
                & w/o Collision\tnote{*}  & 52.74 & 51.58 & 0.0651 & 48.32 & 24.92 & 68.78 \\
                & w/o Joint\tnote{*}  & 47.26 & 46.59 & 0.0701 & 43.40 & 20.22 & 60.79
   \\\bottomrule
    \end{tabular}
    \begin{tablenotes}
    \footnotesize
        \item[*] \blue{The 'w/o' variants represent ablations that omit the guidance of the associated constraint from our method.}
    \end{tablenotes}
\end{threeparttable}
}
\label{tab: joint_proximity_results}
\end{table}

\subsubsection{\blue{Comparison with Other Constraint-Aware Grasp Generation Methods}}
\label{subsec: comparison_other_constraint_aware_generation}

\begin{table}[!h]
\centering
\caption{\blue{Comparison with other representative diffusion-based constraint-aware grasp generation methods (UR5)}}
\resizebox{\linewidth}{!}{
\begin{threeparttable}
\blue{
    \begin{tabular}{c@{\hspace{8pt}}c@{\hspace{2pt}}c@{\hspace{2pt}}c@{\hspace{1pt}}c@{\hspace{1pt}}c@{\hspace{1pt}}c@{\hspace{1pt}}c}
    \toprule
                                     &                                       & \multicolumn{3}{c}{\textbf{C.   Satisfaction(\%)↑}}                        &          \\ \cmidrule(l{0pt}r{0.8em}){3-5}  
    \multirow{-2.5}{*}{\textbf{Scene}} & \multirow{-2.5}{*}{\textbf{Method}}     & CFR                      &  RFR                      & FR                       & \multirow{-2.5}{*}{\begin{tabular}{c} GSR \\ (\%)↑ \end{tabular}}                      &
    \multirow{-2.5}{*}{\begin{tabular}{c} SR \\ (\%)↑ \end{tabular}}    &
    \multirow{-2.5}{*}{\begin{tabular}{c} OSR \\ (\%)↑ \end{tabular}}                      \\ \midrule
                                     & Baseline (PD)  & 77.80 & 99.99 & 77.80 &  40.98 & 31.88 & 87.43 \\
                                     & Baseline (PSR)  & 82.38 & 99.99 & 82.38 & 24.59 & 20.26 & 78.03 \\
    \multirow{-3}{*}{S1}             & \cellcolor[HTML]{D9D9D9}\textbf{Ours} & \cellcolor[HTML]{D9D9D9}\textbf{88.87} & \cellcolor[HTML]{D9D9D9}\textbf{99.99} & \cellcolor[HTML]{D9D9D9}\textbf{88.87} &  \cellcolor[HTML]{D9D9D9}\textbf{53.08} & \cellcolor[HTML]{D9D9D9}\textbf{47.17} & \cellcolor[HTML]{D9D9D9}\textbf{91.77} \\ \midrule
    & Baseline (PD) & 95.03 &  \textbf{99.97} & \textbf{95.01} & 19.37 & 18.40 & 76.74 \\
                 & Baseline (PSR) & \textbf{99.98} & 62.29 & 62.29 & 38.01 & 23.68 & 81.92 \\
    \multirow{-3}{*}{Fig.~\ref{fig: case_study_all} (b)}       & \cellcolor[HTML]{D9D9D9}\textbf{Ours} & \cellcolor[HTML]{D9D9D9}99.25 & \cellcolor[HTML]{D9D9D9}65.58 & \cellcolor[HTML]{D9D9D9}65.57 & \cellcolor[HTML]{D9D9D9}\textbf{53.04} & \cellcolor[HTML]{D9D9D9}\textbf{34.79} & \cellcolor[HTML]{D9D9D9}\textbf{89.42} \\
    \midrule
    \multirow{2}{*}{\textbf{Scene}} &
    \multirow{2}{*}{\textbf{Method}} &
    \multirow{2}{*}{\begin{tabular}{c} CFR \\ (\%)↑ \end{tabular}} &
    \multirow{2}{*}{\begin{tabular}{c} FR \\ (\%)↑ \end{tabular}} &
    \multirow{2}{*}{\begin{tabular}{c} AJP \\ ($\text{rad}^2$)↓ \end{tabular}} &
    \multirow{2}{*}{\begin{tabular}{c} GSR \\ (\%)↑ \end{tabular}} &
    \multirow{2}{*}{\begin{tabular}{c} SR \\ (\%)↑ \end{tabular}} &
    \multirow{2}{*}{\begin{tabular}{c} OSR \\ (\%)↑ \end{tabular}} \\
    & & & & & & & \\
    \midrule
    & Baseline (PD) & 58.92 & 58.86 & 0.0569 & 41.12 & 24.20 & 81.11 \\
                 & Baseline (PSR) & \textbf{82.34} & \textbf{79.68} & 0.0666 & 38.99 & 31.07 & 87.00 \\
    \multirow{-3}{*}{Fig.~\ref{fig: case_study_all} (c)}       & \cellcolor[HTML]{D9D9D9}\textbf{Ours} & \cellcolor[HTML]{D9D9D9}76.45 & \cellcolor[HTML]{D9D9D9}74.76 & \cellcolor[HTML]{D9D9D9}\textbf{0.0398} & \cellcolor[HTML]{D9D9D9}\textbf{49.79} & \cellcolor[HTML]{D9D9D9}\textbf{37.72} & \cellcolor[HTML]{D9D9D9}\textbf{90.26} \\
    \bottomrule
    \end{tabular}
    \begin{tablenotes}
    \footnotesize
        \item[*] \blue{A visualization of the Narrow Corridor scene (S1) is provided in Appendix~A-A.}
    \end{tablenotes}
}
\end{threeparttable}}
\label{tab: baseline_comparison_coll_and_ik_on_ur}
\end{table}

\blue{As shown in Tab.~\ref{tab: baseline_comparison_coll_and_ik_on_ur}, among all constraint-aware grasp generation methods, the proposed approach consistently achieves the highest grasp quality (GSR) and the largest number of successful grasps (SR and OSR). In contrast, PSR achieves the lowest GSR in most cases, except for workspace-boundary grasps (Fig.~\ref{fig: case_study_all}(b)), as it ignores the target grasp distribution during refinement.
PD shows performance closest to Ours, however, 
it performs local optimization on joint samples instead of iteratively selecting the optimal IK solution as (\ref{eq:inner_opt}) in Ours, leading to insufficient exploration of the solution space and a lower GSR. Consequently, to ensure adequate exploration, it must maintain a large batch of initial joint samples, which substantially increases computational cost.
Moreover, the formulation in \cite{ko2025simultaneous} (Eq. (1)) lacks a mechanism to balance reachability with grasp quality. This limitation results in notably lower GSR when grasping near workspace boundaries (i.e., the scene depicted by Fig.~\ref{fig: case_study_all}(b)).
}

\subsubsection{Illustration of Representative Cases}
\label{sec: sim_case_study}
%
For better understanding, we visualize representative cases that illustrate the effect of incorporating various constraints, as depicted in Fig.~\ref{fig: case_study_all}. The denoising process for both our method and the baseline begins with the same Gaussian noise sample. In Fig.~\ref{fig: case_study_all} (a), the collision is progressively resolved with guidance, whereas the baseline leads to significant penetration. In Fig.~\ref{fig: case_study_all} (b), the wrist pose is incrementally projected into the arm's reachable space, whereas the baseline wrist pose remains out of reach. In Fig.~\ref{fig: case_study_all} (c), the fully denoised arm with guidance is closer to the green reference configuration, indicating a shorter joint-space movement from the last grasp attempt.
Among all these cases, the constraint losses converge to zero using our method, while the baseline results in large violations.

\vspace{-10pt}
\subsection{Real-World Experiments}
\label{sec: real_world_exp}

\begin{figure*}
    \centering
    \includegraphics[width=1\linewidth]{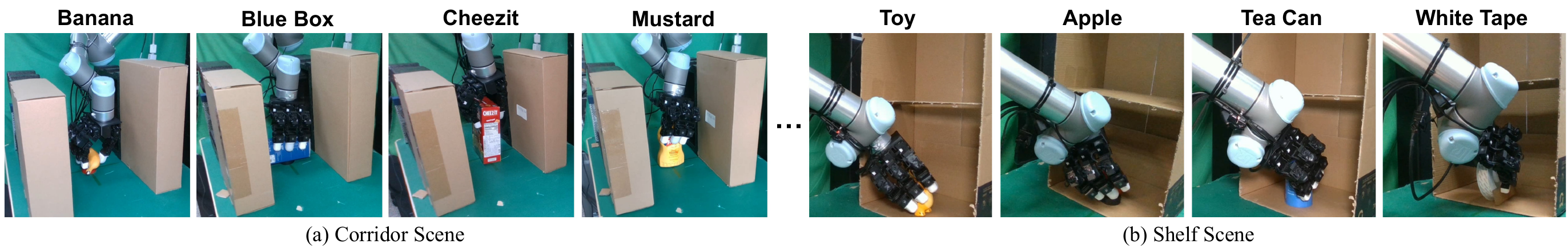}
    \caption{Snapshots of the real-world grasp execution with UR5 and LEAP Hand, under two challenging scenarios. (a) Corridor Scene, (b) Shelf Scene.}
    \label{fig: real_world_collision_snapshots}
    \vspace{-10pt}
\end{figure*}

\begin{table*}[tb]
\centering
\caption{Per-object grasp execution success rate with the grasps generated for eight daily objects in two real-world constrained scenarios (Fig.~\ref{fig: real_world_collision_snapshots})}
\resizebox{0.8\linewidth}{!}{
    \renewcommand{\arraystretch}{1.1}
    \begin{tabular}{cccccccccc}
    \toprule
    \multirow[c]{2}{*}{\textbf{Corridor Scene}} & Object & Banana & Blue Box & Cheezit & Mustard & Toy & Apple & Tea Can & White Tape \\ \cline{2-10}
     & Top-10 Success Num & 8/10 & 6/10 & 8/10 & 8/10 & 9/10 & 10/10 & 7/10 & 10/10 \\
    \midrule
    \multirow[c]{2}{*}{\textbf{Shelf Scene}} & Object & Banana & Blue Box & Cheezit & Mustard & Toy & Apple & Tea Can & White Tape \\ \cline{2-10}
     & Top-10 Success Num & 10/10 & 9/10 & 8/10 & 9/10 & 9/10 & 10/10 & 9/10 & 10/10 \\
    \bottomrule
    \end{tabular}
}
\label{tab: real_collision_grasp_succ_rate}
\vspace{-10pt}
\end{table*}

%
In the real-world experiments, we use the LEAP Hand mounted on a UR5 robotic arm. An Azure Kinect depth camera captures the object's partial point cloud.
We employ cuRobo to plan 
collision-free hand-arm trajectories from the initial configuration to the pre-grasp configuration, where the fingers are slightly spread apart without contacting the object.
\blue{We train a grasp quality evaluator that predicts the probability of successfully executing a given grasp from the object’s partial point cloud and grasp pose, allowing selection of the high-quality candidates for real-world execution (see Appendix C for more details). The arm configuration is computed using the same IK selection strategy as described in the Evaluation Metrics section. More real-world experiment results can be found in Appendix B.}
%

%
%

%
\begin{figure}
    \centering
    \includegraphics[width=1\linewidth]{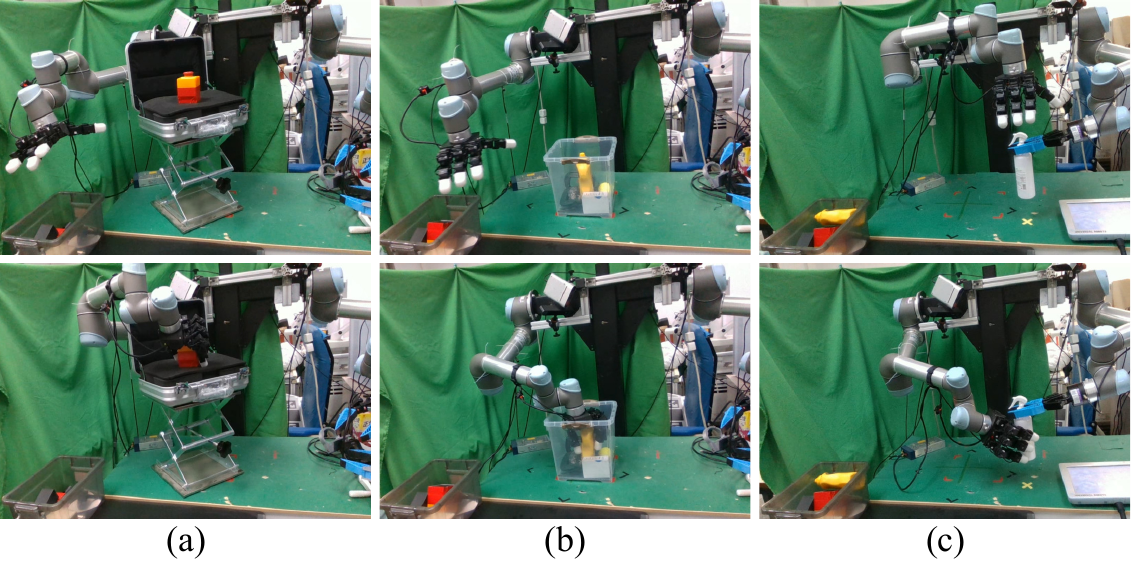}
    \caption{\blue{Snapshots of real-world constrained grasping demonstrating the practical effectiveness of the proposed method. Top: the initial configurations. Bottom: the final grasping configurations. (a) Grasping behind obstacles. (b) Grasping in deep containers. (c) Bi-manual handover grasping.}}
    \label{fig: real_snapshots_practical}
    \vspace{-10pt}
\end{figure}

\subsubsection{Evaluation of Grasp Quality in the Real-World}
%
We first evaluated the quality of the generated grasps in the real world, which is necessary as simulations cannot accurately capture the contact-rich interactions present in the physical world.
The evaluation involved grasping eight everyday objects with varying geometries and masses, positioned in two challenging scenarios: (a) Corridor Scene and (b) Shelf Scene.
For each object, we sampled 40 candidate grasps in a batch and executed the top 10 arm-feasible grasps with the highest predicted success probabilities.
Fig.~\ref{fig: real_world_collision_snapshots} shows snapshots of executed grasp poses, while Tab.~\ref{tab: real_collision_grasp_succ_rate} highlights the success rates, demonstrating reasonable performance in challenging real-world scenarios. Further discussion of failure cases can be found in Appendix~B.

\subsubsection{\blue{Demonstration of Constrained Grasping Highlighting Practical Utility}}
\blue{We validate the practical effectiveness of the proposed method in several challenging real-world setups where rejection sampling often fails. As shown in Fig.~\ref{fig: real_snapshots_practical}, our method enables (a) downward-facing grasps in deep containers, (b) side grasps for objects behind obstacles under collision and reachability constraints, and (c) collision-aware bi-manual handover with minimal joint motion.}

\balance
\section{Conclusions}
\textbf{Conclusion}: In this work, we present an arm-aware dexterous grasp generation framework that integrates arm kinematics and environmental constraints into diffusion-based sampling while retaining the generalization ability of pretrained arm-agnostic grasp models. By jointly optimizing wrist poses and arm configurations and injecting closed-form arm-related gradients during denoising, the method guides sampling toward feasible regions. Experiments in simulation and real-world settings featuring constrained scenarios show improved constraint satisfaction and arm configuration proximity, without requiring arm-specific data synthesis and training, while maintaining reasonable success rates across different arms and environments. 

\textbf{Discussion}: 
We note that the incorporation of guidance may slightly perturb the learned distribution and affect grasp quality. To address this, null-space projection and evaluator-based gradient guidance can be introduced during denoising to maintain alignment with high-quality grasp regions. Moreover, as the refinement induced by the guidance is inherently local, the method is most effective at converting near-feasible samples into feasible ones, motivating the use of multiple initial samples for broader feasible coverage. Future work will explore integrating our guidance with improved initial sampling strategies to further enhance global performance.
 
{\small
\bibliography{IEEEabrv,ref}
\bibliographystyle{IEEEtran}
}

\newpage
{
\appendices
\renewcommand{\thesection}{\Roman{section}}

\section{Additional Results and Analysis}
\label{sec: additional_resuls_and_analysis}

\subsection{Design and Visualization of the Evaluation Scenes}
\label{sec: design_vis_evaluation_scenes}

\begin{figure*}[!ht]
    \centering
    \begin{subfigure}[b]{0.24\linewidth}
        \centering
        \includegraphics[width=\linewidth]{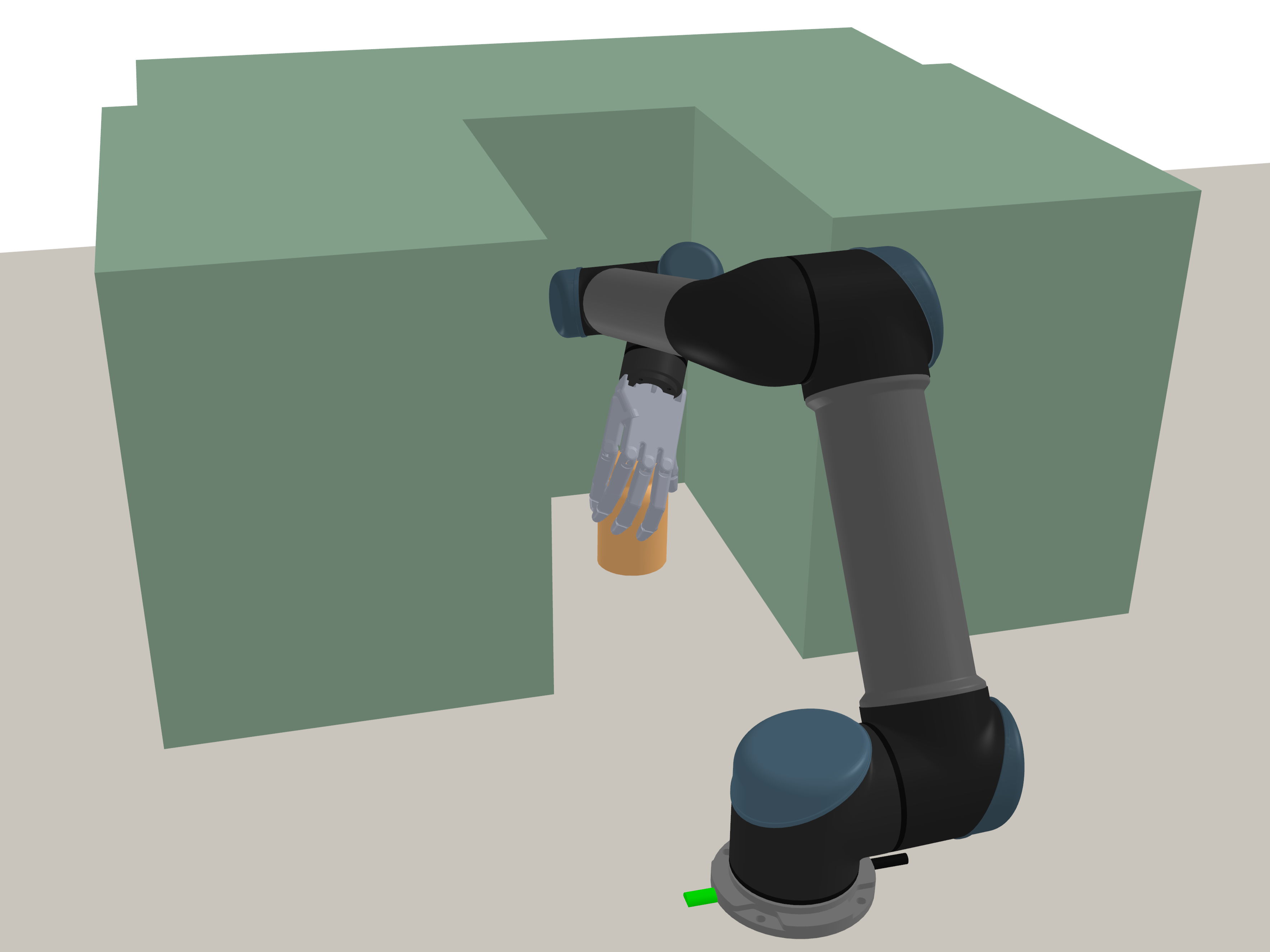}
        \caption{Narrow Corridor (S1)}
        \label{fig: narrow_corridor_scene}
    \end{subfigure}
    \hfill
    \begin{subfigure}[b]{0.24\linewidth}
        \centering
        \includegraphics[width=\linewidth]{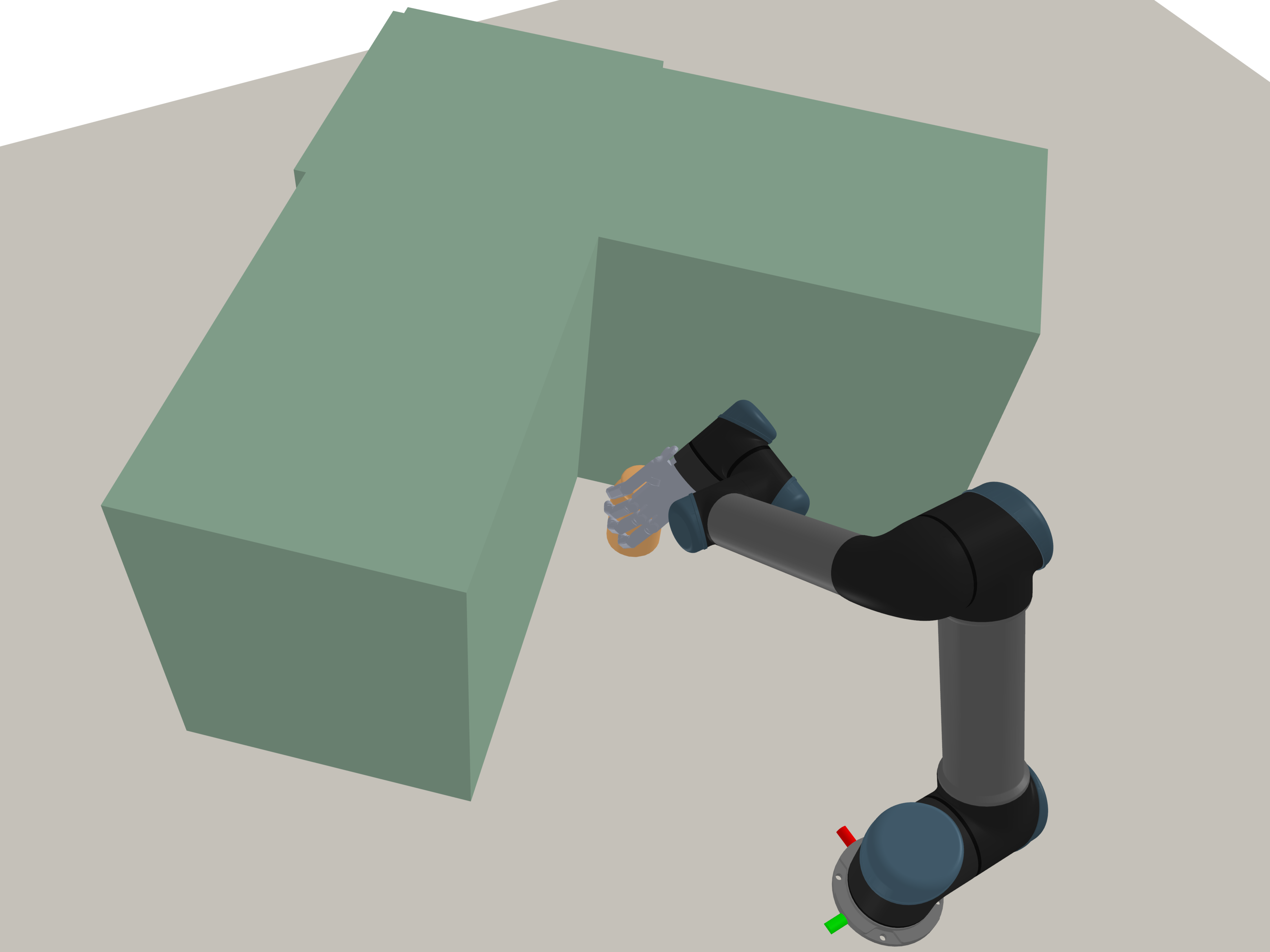}
        \caption{Room Corner (S2)}
        \label{fig: room_corner_scene}
    \end{subfigure}
    \hfill
    \begin{subfigure}[b]{0.24\linewidth}
        \centering
        \includegraphics[width=\linewidth]{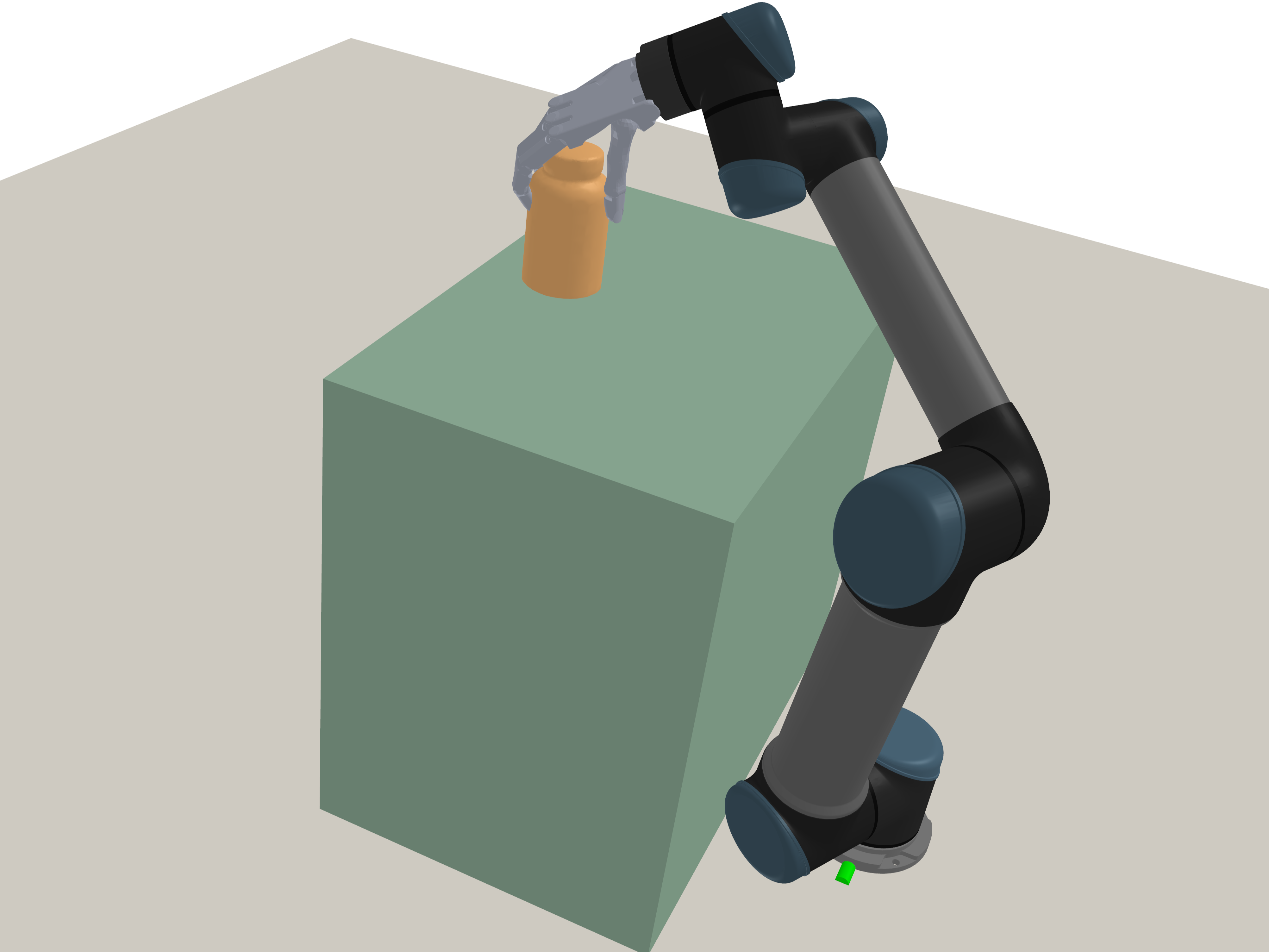}
        \caption{Shelf Top (S3)}
        \label{fig: shelf_top_scene}
    \end{subfigure}
    \hfill
    \begin{subfigure}[b]{0.24\linewidth}
        \centering
        \includegraphics[width=\linewidth]{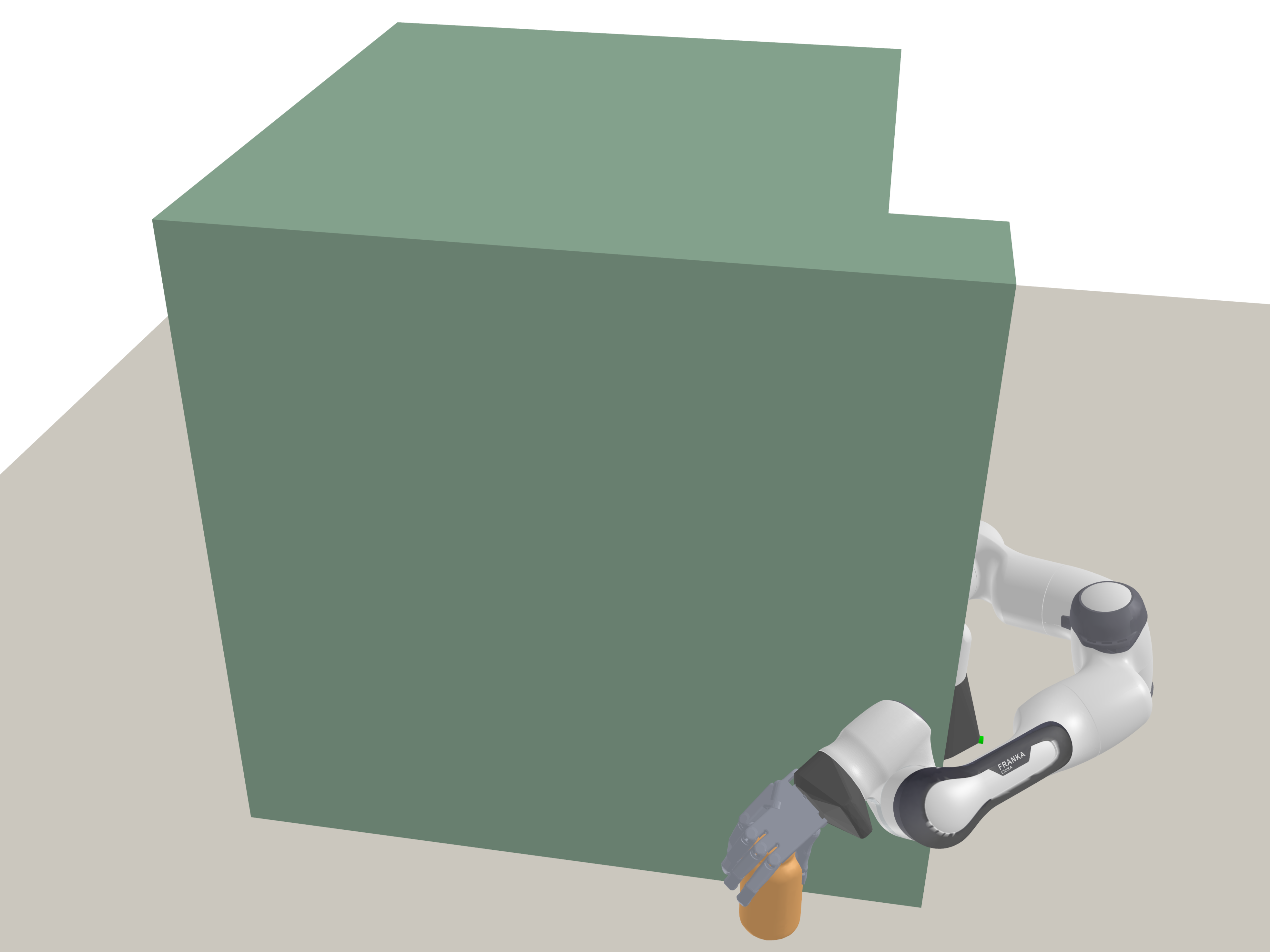}
        \caption{Partition Board (S4)}
        \label{fig: partition_board_scene}
    \end{subfigure}

    \caption{Visualization of four evaluation scenarios (S1-S4) with one successful grasp involving the robotic arm and dexterous hand, the obstacles painted in green, and the grasped object highlighted in orange. Scene S4 is specifically designed to showcase Franka's collision avoidance using its kinematic redundancy.}
    \label{fig: visualize_four_scenes_collision}
\end{figure*}
To evaluate the proposed method, we designed six challenging scenarios featuring high obstacle density and grasps near the workspace boundary. Four scenarios (S1–S4) focusing on collision avoidance are visualized in Fig.~\ref{fig: visualize_four_scenes_collision}. The remaining two, targeting hand reachability and joint proximity constraints, are illustrated in Fig.~\ref{fig: case_study_all}(b) and (c) in the main text, respectively.

\subsection{Parameter Sensitivity Analysis}
\label{sec: sensitivity_analysis}
We have conducted systematic sensitivity analyses on key hyperparameters in our method. All experiments are performed on a randomly sampled subset of DexGraspNet (DGN), consisting of 2,308 objects with various geometries, poses, and scales. 
We analyze the effects of the constraint weight $\lambda_c$ and the learning rate $\eta_\lambda$ under the collision-avoidance constraint in the Narrow Corridor scenario (Fig.~\ref{fig: visualize_four_scenes_collision} (a)). Following common practice in guided diffusion sampling \cite{Liu2021MoreCF}, we apply magnitude clipping to the pose gradients to prevent numerical singularities and divergence under large guidance strength. The heatmap results (Fig. \ref{fig: collision_sweep_3}) reveal consistent trends across the two arms, 1) FR and OSR increase rapidly with larger $\lambda_c$ and $\eta_\lambda$ and quickly enter a saturation regime at moderate-to-high values; 2) GSR exhibits only mild degradation during this process, indicating limited sensitivity to the guidance parameters; and 3) The learning rate affects mainly the convergence speed of OSR rather than the final performance.

\begin{figure}[!b]
    \centering
    \includegraphics[width=\linewidth]{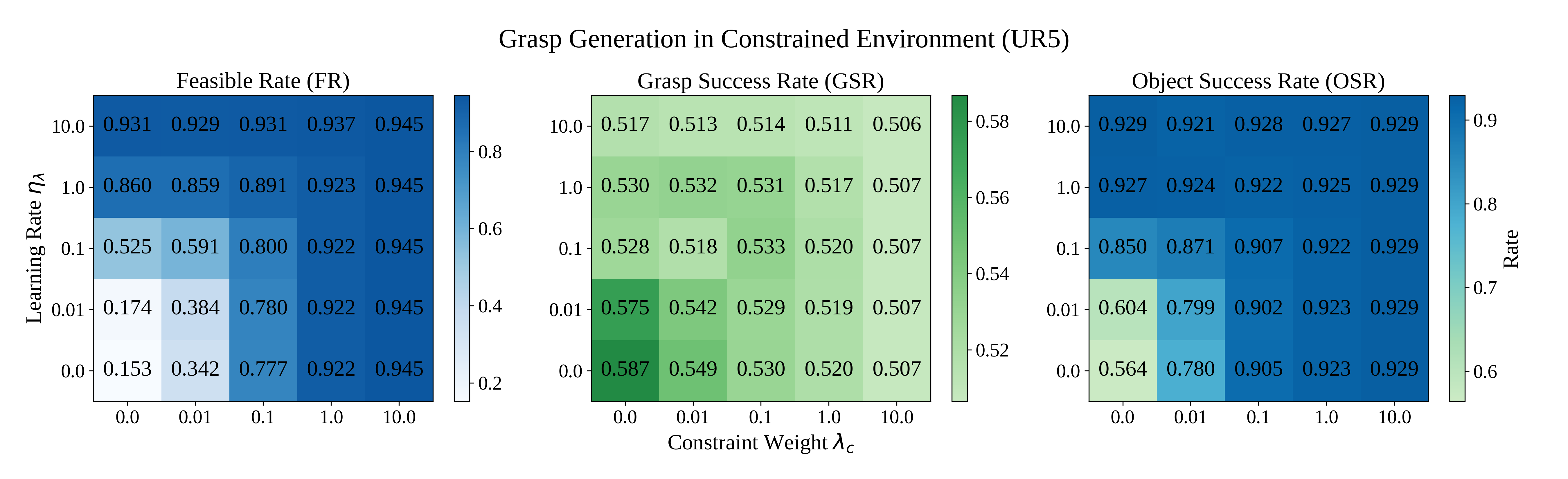}
    \caption{Effect of different constraint weights $\lambda_c$ and learning rates $\eta_\lambda$. }
    \label{fig: collision_sweep_3}
\end{figure}

\begin{figure}[!b]
    \centering
    \includegraphics[width=0.7\linewidth]{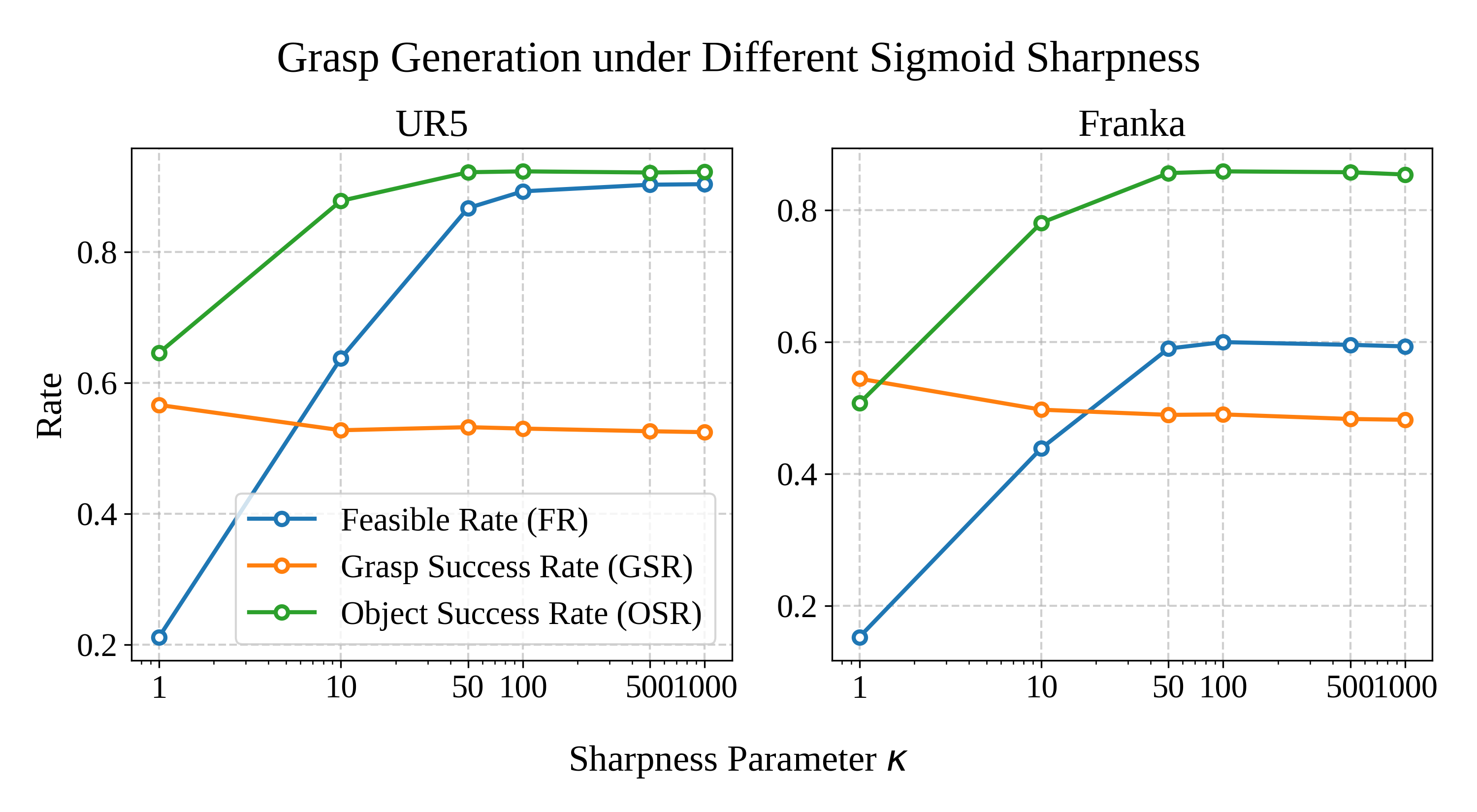}
    \caption{Effect of different sigmoid shape.} 
    \label{fig: sigmoid_sharp_3}  
\end{figure}

The result of the effect of the sigmoid shape (Fig. \ref{fig: sigmoid_sharp_3}) shows that when the sharpness parameter $\kappa > 50$, all metrics converge and remain stable. This behavior can be attributed to the fact that the signed distance field (SDF) values are typically on the order of $10^{-2}\sim 10^{-1}$ in the real environment, for which $\kappa$ in the range of $50 \sim 500$ provides an appropriate scale to capture the variations of the SDF, leading to stable optimization.

\begin{figure}[!b]
    \vspace{-5pt}
    \centering
    \hspace*{-0.07\linewidth} 
    \includegraphics[width=1.1\linewidth]{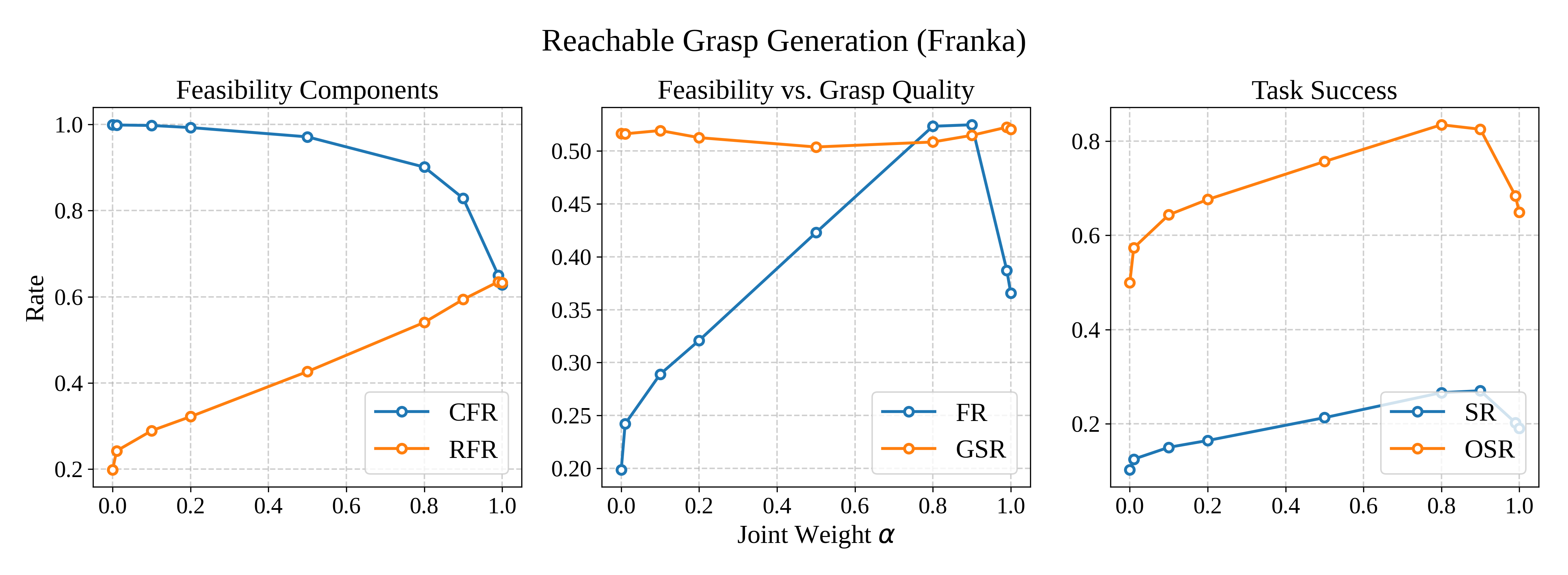}
    \vspace{-5pt}
    \caption{Effect of different weights coefficient $\alpha$.}
    \label{fig: reach_sweep_3}
    \vspace{-5pt}
\end{figure}

In the reachable grasp generation experiment, we analyze the weight coefficient $\alpha$ that balances collision-avoidance and reachability (Fig. \ref{fig: reach_sweep_3}). Results under a linear sweep of $\alpha$ in [0,1] reveal stable trends across both arms: 1) RFR increases monotonically with $\alpha$, while CFR exhibits a slight decrease, indicating a clear trade-off between reachability and collision avoidance; 2) FR and GSR demonstrate a balanced relationship, with GSR showing only mild sensitivity to $\alpha$, reflecting the robustness of the proposed guidance method.

In the grasp generation experiment with reference joint configuration, we analyze the effect of the joint regularization weight $\lambda_s$. As shown in Fig. \ref{fig: joint_sweep_3}, the results reveal a clear trade-off between joint proximity and task success. Specifically, AJP decreases monotonically with increasing $\lambda_s$, indicating that the generated grasps remain closer to the reference joint configuration. However, when $\lambda_s$ exceeds a certain threshold, the grasp gradually deviates from object-centric configurations, leading to a rapid drop in GSR and OSR. This behavior suggests that joint regularization acts as a relatively strong constraint. 

\begin{figure}[!htb]
    \vspace{-5pt}
    \centering
    \hspace*{-0.07\linewidth} 
    \includegraphics[width=1.1\linewidth]{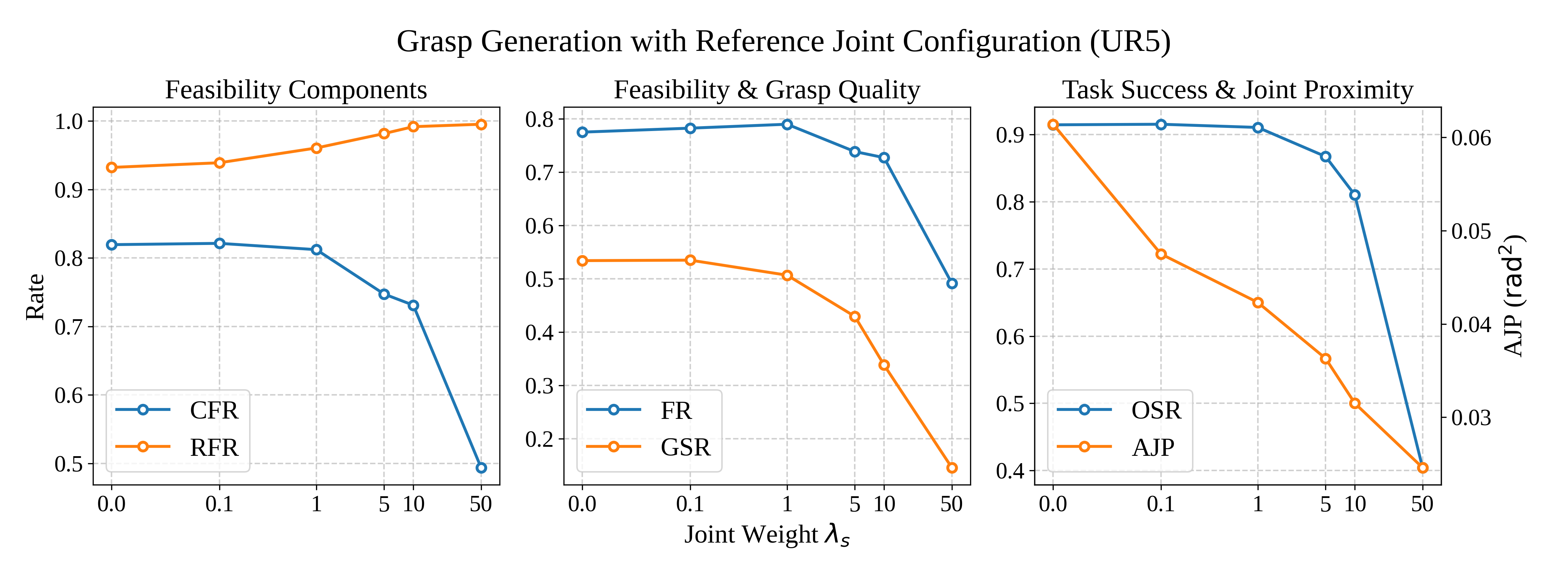}
    \vspace{-5pt}
    \caption{Effect of different joint weights $\lambda_s$. }
    \label{fig: joint_sweep_3}
    \vspace{-5pt}
\end{figure}

We finally evaluate the effect of the number of diffusion denoising steps $\text{T}_{\text{N}}$ during sampling. Experiments are conducted in the Narrow Corridor scenario, with $\text{T}_{\text{N}} = 5, 10, 15, 20$ (Fig. \ref{fig: time_step_sweep_3}). The results show that using a very small number of denoising steps (e.g, 5) leads to noticeable degradation across all metrics. When $\text{T}_{\text{N}} = 10$, OSR already approaches its maximum, indicating that the proposed DDIM-based guided diffusion achieves stable performance with few steps. Further increase $\text{T}_{\text{N}}$ allows for more fine-grained pose refinement, which can improve FR, but may slightly reduce GSR. Considering both sampling efficiency and task success, we adopt $\text{T}_{\text{N}} = 10$ as the default setting in our experiments.

\begin{figure}[!htb]
    \vspace{-5pt}
    \centering
    \includegraphics[width=0.7\linewidth]{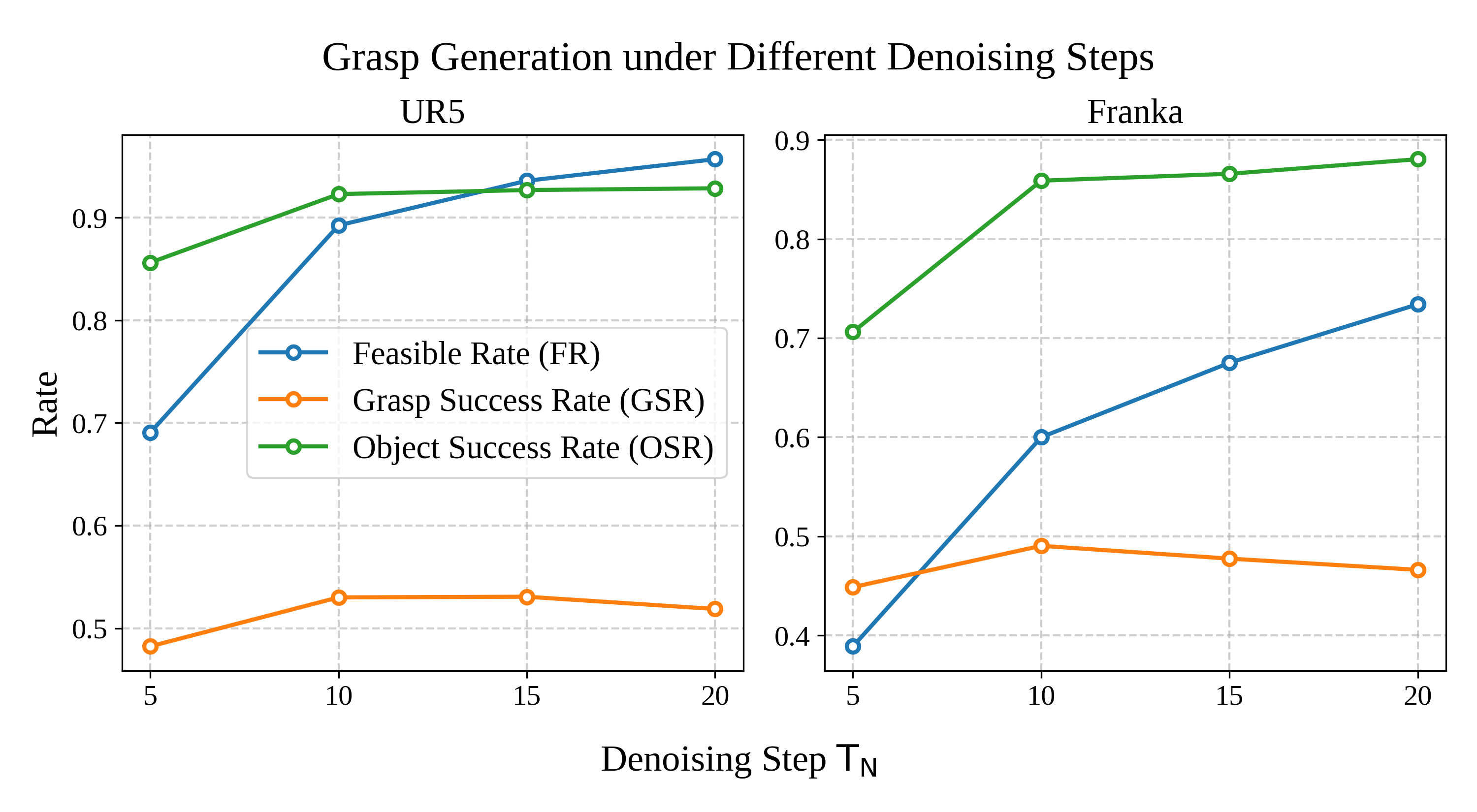}
    \hfill
    \caption{Effect of different denoising step $\text{T}_{\text{N}}$.}
    \label{fig: time_step_sweep_3}  
    \vspace{-5pt}
\end{figure}






%
%

%
\subsection{Ablation of the Nullspace Projection Method and the Noised Gradient}
\label{sec: ablation_nullspace_proj}
We propose a nullspace projection method to address the potential decline in grasp quality caused by the introduction of guidance, which ultimately maintains grasp quality. The main task is defined as keeping a point on the palm fixed in the world coordinates. By projecting the guidance gradient $\nabla g$ to the main task's nullspace, the hand keeps facing the object in most cases, thus maintaining the grasp quality. The ablation results in Tab.~\ref{tab: ablation_proj_and_noised_grad} show the percentage change of metrics when the nullspace projection is enabled (w/ proj.) compared to '\textbf{Ours}' in Tab.~\ref{tab: collision_graspgen_result_ur5} of the main text. As shown by the results, the GSR increases with projection, indicating improved grasp quality. However, there is minimal change in the SR, indicating that the absolute number of successful grasps in a batch remains consistent. This trend arises from the interplay between the decrease in feasible grasps (FR) and the increase in grasp quality (GSR). The decline in FR is expected, as the direction of gradient descent is modified by the projection. The OSR presents similar trends to SR.

\begin{table}[!htb]
\centering
\caption{Performance of the ablation groups compared to '\textbf{Ours}' in Tab.~\ref{tab: collision_graspgen_result_ur5}}
\resizebox{0.9\linewidth}{!}{
    \begin{tabular}{cccccc}
    \toprule
    \multirow{2.5}{*}{\textbf{Scene}} & \multirow{2.5}{*}{\textbf{Method}} & \multicolumn{4}{c}{\textbf{Changes of Metrics(\%)↑}} \\ \cmidrule(l){3-6} 
                                    &                                  & FR      & GSR      & SR      & OSR      \\ \midrule
    \multirow{2}{*}{S1}             & w/ proj. & - 4.80 & + 2.54 & - 0.41 & + 0.34 \\
                                    & w/ noised grad. & - 0.75 & - 3.69 & - 3.64 & - 0.12 \\ \midrule
    \multirow{2}{*}{S2}             & w/ proj. & - 18.39 & + 5.52 & - 5.19 & - 1.41 \\
                                    & w/ noised grad. & - 0.12 & - 1.96 & - 2.29 & - 0.11 \\ \midrule
    \multirow{2}{*}{S3}             & w/ proj. & + 0.83 & + 0.81 & + 0.94 & + 0.33 \\
                                    & w/ noised grad. & + 10.93 & - 0.70 & + 5.07 & + 2.48 \\ \bottomrule
    \end{tabular}
}
\label{tab: ablation_proj_and_noised_grad}
\end{table}

%
When the noised wrist pose $\bm{x}_t$ instead of $\hat{\bm{x}}_0$ is used to compute $\nabla g$ in (\ref{eq: primal_update_constrained_sampling}), we assume that the resulting noised gradient may result in poor convergence. The ablation results with noised gradient (w/ noised grad.) are presented in Tab.~\ref{tab: ablation_proj_and_noised_grad}. The FR decreased by 0.75\% and 0.12\% in S1 and S2 as expected, while it increased by 10.93\% in S3. The obstacles of S1 and S2 result in the robotic arm alternately collides with two walls and produces inconsistent gradients which hinder convergence. While for S3 with fewer obstacles, $\nabla g(\bm{x}_t)$ performs better by providing more direct and timely guidance. Therefore, in specific applications, the choice between $\bm{x}_t$ and $\hat{\bm{x}}_0$ should be determined by the metrics relevant to the particular scenario. In addition, the grasp quality marginally decreases (GSR), and the change of SR and OSR depends on the specific scene.
\subsection{Computation Time Cost}
\label{sec: compuation_time_cost}
We report inference-time cost measured as the average inference time per successful grasp. Fig. \ref{fig: average_cost_2} shows the average inference time evaluated over 2,308 objects, with 10 candidate grasps generated for each object, under different batch sizes on UR5 and Franka. All experiments are conducted on a single NVIDIA RTX 3090 Ti GPU. Across all settings, the proposed method generates one successful grasp within approximately 0.5–2 ms on average. Although gradient-related computations are evaluated at each diffusion step, the inference time per successful grasp remains lower than that of the rejection-based baseline under the same number of denoising steps. This behavior is mainly attributed to the improved feasibility and success rates introduced by guidance, which reduce the number of discarded samples. In addition, inference time decreases with increasing batch size, indicating that the guidance-related computations are efficiently processed in batches.

\begin{figure}[!htb]
    \centering
    \includegraphics[width=1\linewidth]{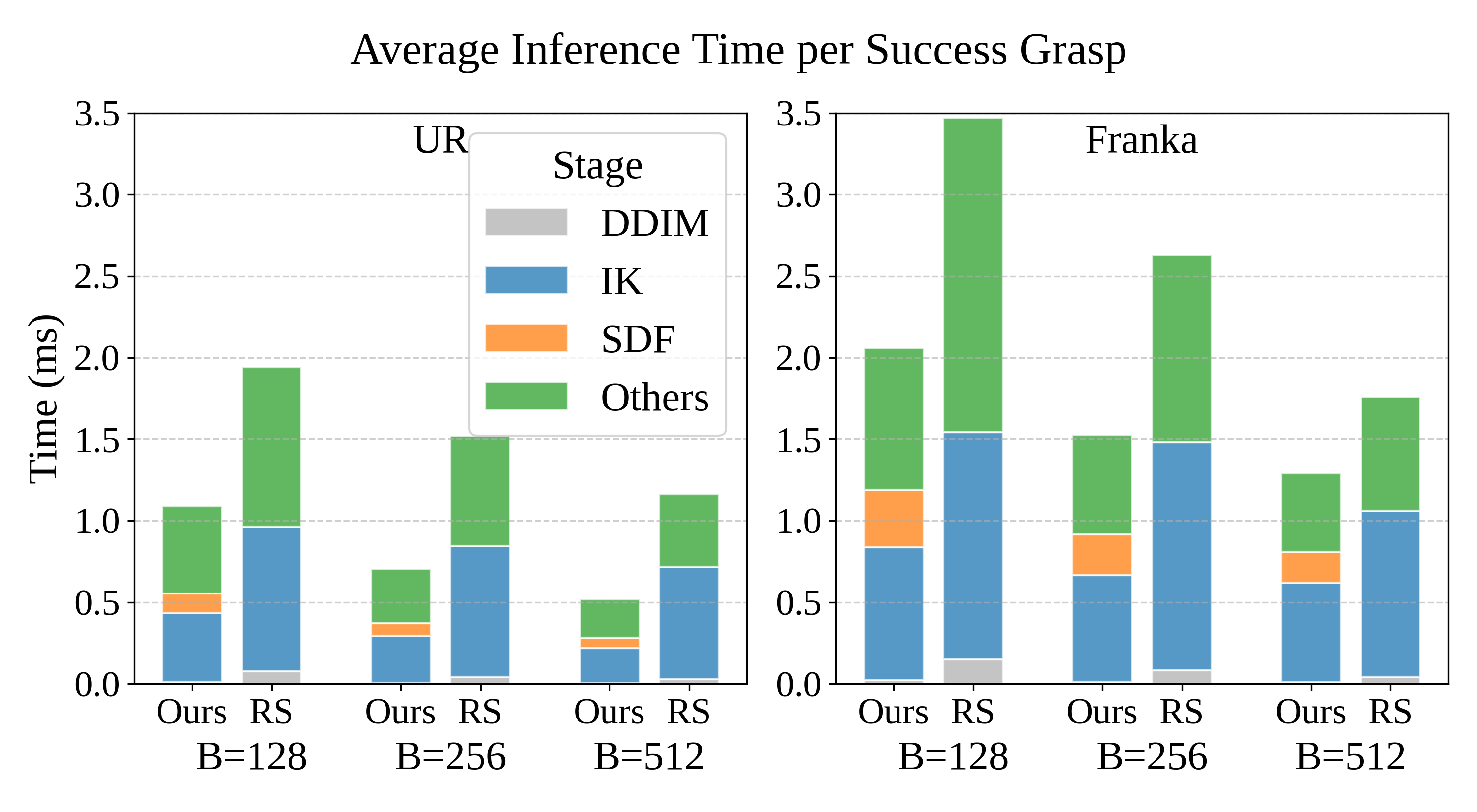}
    \vspace{-5pt}
    \caption{The average inference time per successful grasp under different batch size on UR5 and Franka in the Narrow Corridor scenario.}
    \label{fig: average_cost_2}  
    \vspace{-5pt}
\end{figure}

The inference time is decomposed into SDF, IK, and other components in Tab. \ref{tab: time_breakdown}. The SDF query constitutes the dominant portion of the runtime, as it involves evaluating signed distance values for a large number of sampled points on the robot against each object in the scene. Within the IK component, the computation time is comparably distributed among solver invocation, forward kinematics and Jacobian computation, and error and gradient evaluation, due to repeated calls to the analytical solver and large-scale matrix operations. The remaining computation cost primarily arises from conversions between Lie algebra and Lie group representations.

\begin{table}[!htb]
\centering
\caption{Inference-Time Breakdown of Major Components}
\label{tab: time_breakdown}
\begin{threeparttable}
\begin{tabular}{l l c}
\toprule
\textbf{Module} & \textbf{Component} & \textbf{Time Ratio (\%)} \\
\midrule
\multirow{5}{*}{SDF}
 & Query SDF     & 56.8 \\
 & Link Jacobian & 17.8 \\
 & Sphere Jacobian & 12.2 \\
 & Nullspace Projection & 6.3 \\
 & Others        & 6.9 \\
\midrule
\multirow{4}{*}{IK}
 & Call Solver   & 41.6 \\
 & FK \& Jacobian & 33.4 \\
 & Error \& Gradient & 24.5 \\
 & Others        & 0.6 \\
\midrule
\multirow{4}{*}{Others}
 & Gradient Conversion & 57.4 \\
 & Point Cloud Encoding & 11.5 \\
 & Predict Joints & 5.2 \\
 & Others        & 25.9 \\
\bottomrule
\end{tabular}
\end{threeparttable}
\vspace{-10pt}
\end{table}

\subsection{Influence of the Number of IK Solutions}
\label{sec: influence_num_IK_sols}
Note that solving (\ref{eq: KL_based_constrained_sampling}) involves minimization over the IK solution set $\mathcal{Q}$, which is approximated with a finite set in practice. As shown in Tab.~\ref{tab: ablation_num_ik_solutions}, increasing the number of IK solutions enhances all metrics, indicating that a better solution to (\ref{eq: KL_based_constrained_sampling}) has been found. Note that using fewer than 4 IK solutions can significantly degrade performance with the UR5, as these solutions reside in disconnected sub-manifolds of the configuration space. Their varying order can result in unstable gradients due to the discontinuity of the IK solutions. In contrast, the performance with Franka is less affected, as it has a connected IK solution set. In addition, more inverse kinematics (IK) solutions reduce the number of grasps generated per second. Thus, an appropriate number—8 for UR5 and 10 for Franka—is chosen.
\begin{table}[!b]
 \vspace{-3pt}
\centering
\caption{Performance of grasp generation with varying number of IK solutions}
\resizebox{0.8\linewidth}{!}{
    \begin{tabular}{cccccc}
    \toprule
    \multirow{2.5}{*}{\textbf{Arm}} & \multirow{2.5}{*}{\textbf{IK Num.}} & \multicolumn{4}{c}{\textbf{Metrics(\%)↑}} \\ 
    \cmidrule(lr{0pt}){3-6}
                                  &                                   & FR      & GSR      & SR      & OSR      \\ 
    \midrule
    \multirow{3}{*}{UR5}          
                                  & 2  & 2.74  & 44.01 & 1.21  & 10.06 \\
                                  & 4  & 45.36 & 56.64 & 25.69 & 83.64 \\
                                  & 8  & 73.16 & 57.80 & 42.29 & 90.32 \\ 
    \midrule
    \multirow{3}{*}{Franka}       
                                  & 5  & 35.47 & 48.79 & 17.31 & 71.58 \\
                                  & 10 & 46.18 & 51.78 & 23.91 & 80.46 \\
                                  & 20 & 60.70 & 53.65 & 32.57 & 85.72 \\ 
    \bottomrule
    \end{tabular}
}
\label{tab: ablation_num_ik_solutions}
\end{table}

\section{Additional real-world experiments}
\subsection{Generating Reachable Grasps Near Workspace Boundaries}
We showcased the effectiveness of the proposed method in generating reachable grasps near the arm's workspace boundary. We executed a generated grasp both with and without guidance, as shown in Fig.~\ref{fig: real_ik_exp_snapshots}. Without guidance, the system tends to produce unreachable grasps behind the object, resulting in failures even when the arm is fully extended.
\begin{figure}[!htb]
    \centering
    \includegraphics[width=1\linewidth]{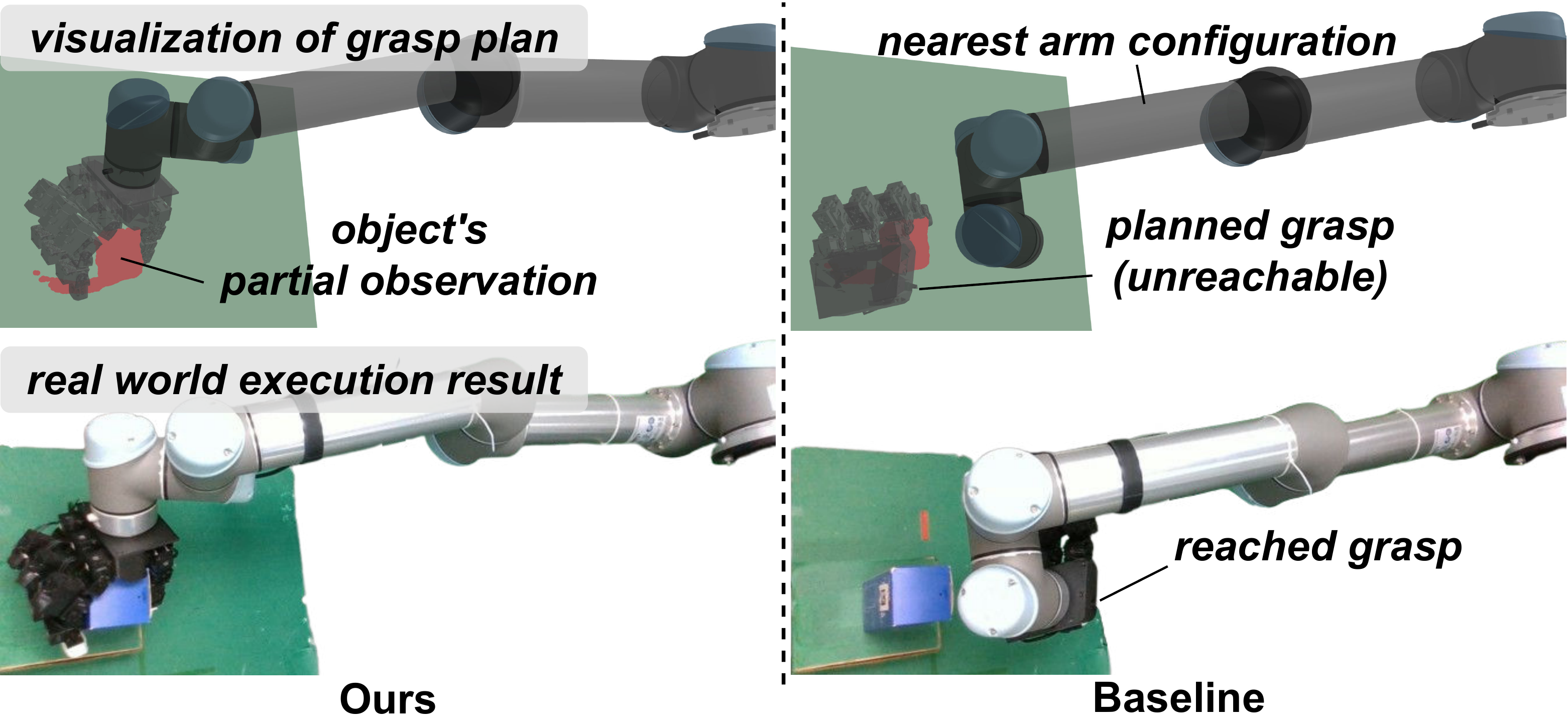}
    \caption{Comparison of grasp generation near the arm's workspace boundary. The top right figure shows a planned grasp that the robotic arm cannot reach.
    }
    \label{fig: real_ik_exp_snapshots}
    \vspace{-13pt}    
\end{figure}

\subsection{Generating Proximal Grasp Configurations for Regrasping}
We demonstrated the proposed method's ability to generate grasps with adjacent arm configurations in joint space during regrasping. Using the initial grasp's arm configuration as a reference, we consecutively sampled grasps 10 times, selecting the one with the lowest joint proximity loss from 40 candidates per sample for execution, as shown in Fig.~\ref{fig: real_joint_exp_snapshots}.
Without guidance refinement, the baseline method leads to unnecessary arm movements.
In contrast, our approach reduces the average joint space distance between grasps from 0.88 rad to 0.56 rad, resulting in a smoother execution trajectory. 

\begin{figure}[!htb]
    \centering
    \includegraphics[width=1\linewidth]{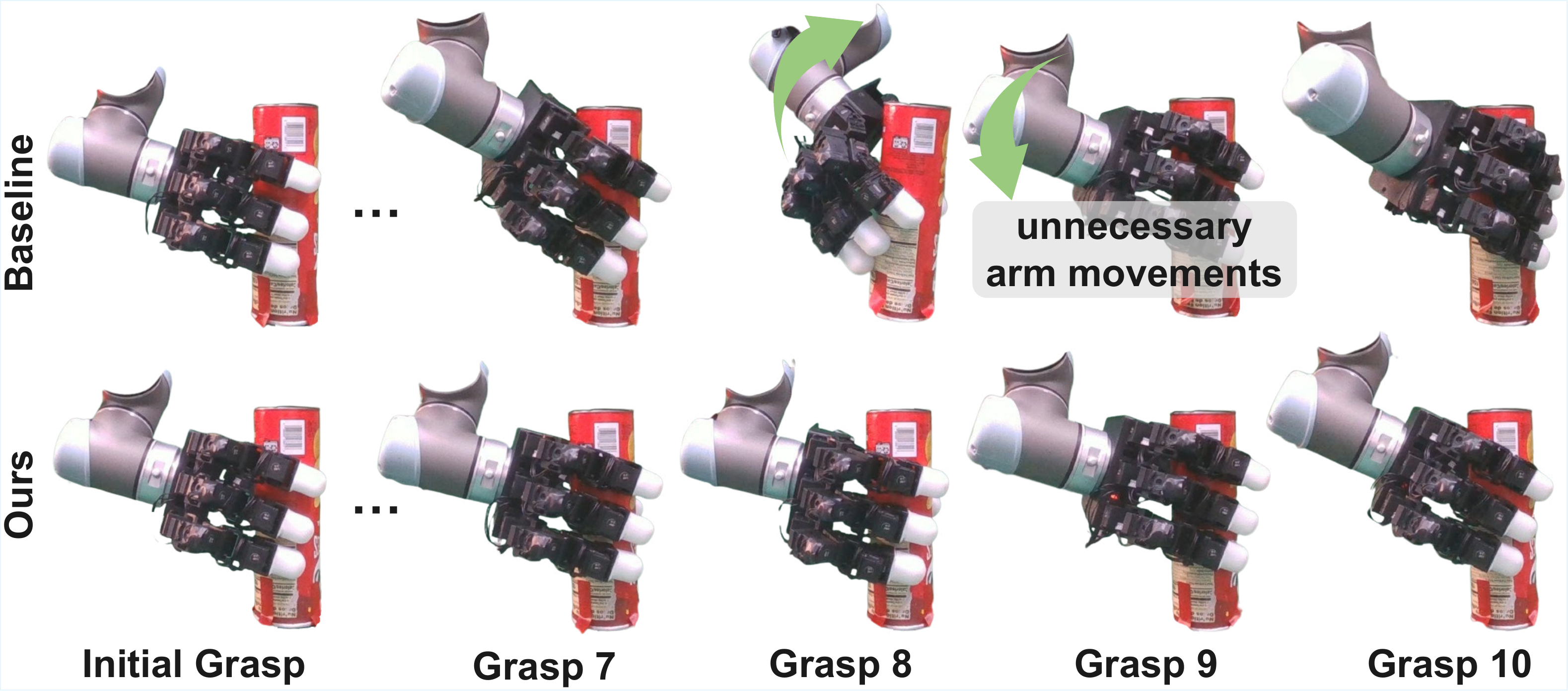}
    \caption{Illustration of guided grasp generation for continuously regrasping with the Joint Proximity constraint. The reference initial grasp and the four final grasps are displayed from left to right, with additional grasps omitted for brevity. Please refer to the attached video for the complete grasp sequence.}
    \label{fig: real_joint_exp_snapshots}
    \vspace{-13pt}
\end{figure}

\subsection{Failure Cases of Executing Generated Grasps in the Real World}
\label{sec: real_exp_failure_cases}
We present several failure cases in Fig.~\ref{fig: failure_case_real}, highlighting potential limitations of the proposed method and possible directions for improvement. 1) In some cases, the guidance compromises the quality of a small subset of generated grasps, resulting in unstable configurations that only grasp the edge of the object (Fig.~\ref{fig: failure_case_real} (a-b)). This accounts for 7 out of 20 failed grasps. This issue often arises when the arm remains in collision until the end of the denoising process, causing conflicts between constraint satisfaction and convergence to the learned grasp distribution.
Although we utilize the grasp evaluator to filter out high-quality grasps, we find that real-world point clouds are often noisier and more incomplete, leading to misestimation of grasp success probabilities. This issue can be mitigated through increased data augmentation and fine-tuning with real-world data.
Another reason for failure is inaccuracies in motion planning and open-loop execution, which can result in unintended collisions, contributing to 9 out of 20 failed grasps (Fig.~\ref{fig: failure_case_real} (c)). This issue can be mitigated through real-time feedback and online sensing. Additionally, incorrect estimation of the object's mass parameters or insufficient friction accounts for the remaining 4 out of 20 failed grasps (Fig.~\ref{fig: failure_case_real} (d)), though this aspect lies beyond the scope of this work. 

\begin{figure}[!htb]
    \centering
    \includegraphics[width=1\linewidth]{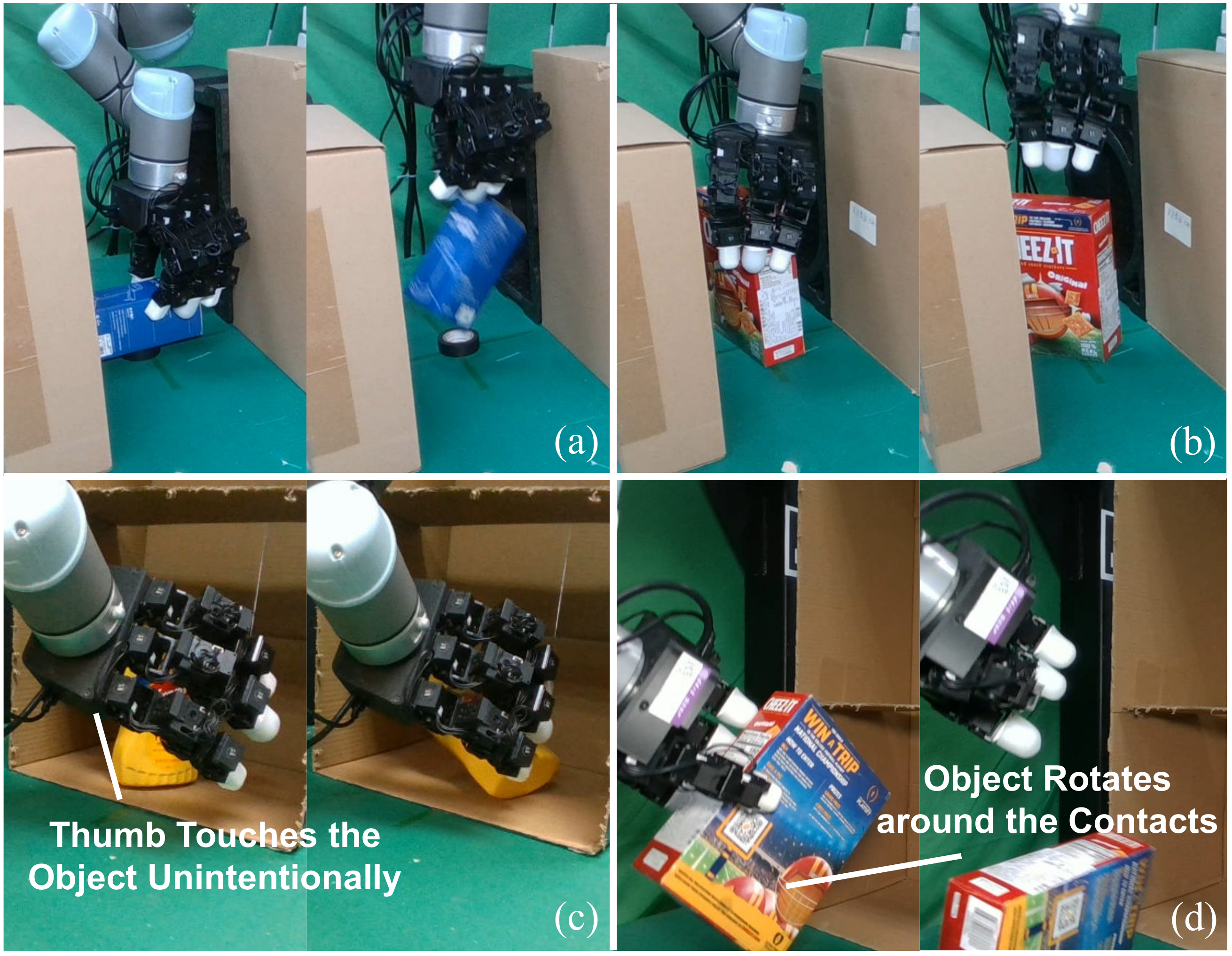}
    \caption{Failure cases in real-world experiments. (a–b) The quality of a small subset of generated grasps is degraded by the guidance, leading to unstable grasps holding only the object’s edge. (c) The dexterous hand unintentionally contacts the object due to open-loop execution. (d) The object shifts in hand because of inaccurate mass parameter estimation or insufficient friction.}
    \label{fig: failure_case_real}

\end{figure}

\section{Additional Details of the Proposed Method}
\begin{figure}[!htb]
    \centering
    \includegraphics[width=1\linewidth]{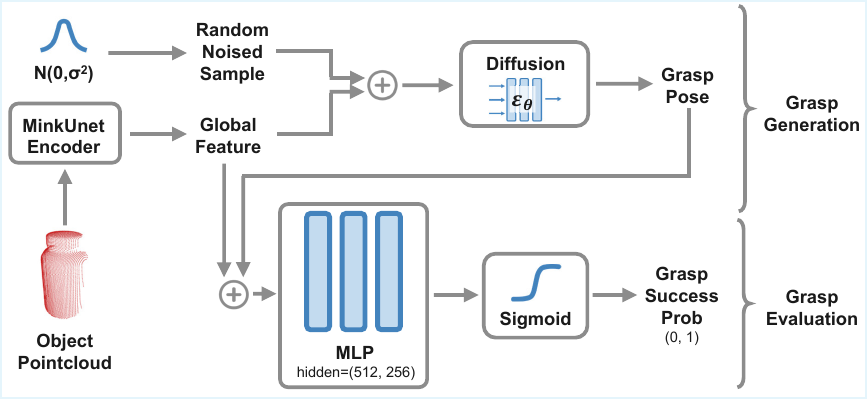}
    \caption{Architecture of the grasp generation and evaluation networks.}
    \label{fig: network_arch}
\end{figure}
\subsection{Network Architecture}
Our network design is inspired by prior work \cite{Chen2024BODexSA}. For grasp pose generation, the object's partial point cloud is embedded into a 1024-dimensional global feature vector using a Minkowski Unet. This global feature, along with a randomly sampled noise from a Gaussian distribution, is input into the denoising process. The noise prediction network, conditioned on the global feature and the time step $t$, predicts a denoised sample from the noisy input. The grasp pose is then recovered from the fully denoised sample and the global feature. A grasp evaluator predicts the probability of successfully executing a given grasp, based on the grasp pose and the global feature. The evaluator is implemented as an MLP with hidden dimensions of $[512, 256]$, followed by a sigmoid activation function. The grasp evaluation uses the same training data as the generation network, augmented with execution success labels from performing all grasps in the MuJoCo simulator. The evaluator is trained using cross-entropy loss and successfully predicts 87\% of grasps in the test set with its best checkpoint.

}

\end{document}